\documentclass[authoryear,final,5p,times]{elsarticle}

\usepackage{amssymb}
\usepackage{makecell}
\usepackage{booktabs}
\usepackage{moresize}

\usepackage{amsmath}
\usepackage{hyperref}

\usepackage{hyphenat}
\hypersetup{
  colorlinks=true,
  linkcolor=red,
  filecolor=magenta,
  urlcolor=cyan,
  citecolor=blue
}

\usepackage{multirow}
\usepackage{cuted}
\usepackage{tocloft}
\usepackage[utf8]{inputenc}
\usepackage{setspace}
\usepackage{graphicx}
\usepackage{subcaption}
\usepackage{tabularx}
\usepackage{tikz}
\usepackage[parfill]{parskip}
\usepackage{setspace}
\usepackage{color,soul}

\journal{Ecological Informatics}

\begin{document}

\begin{frontmatter}



\title{Forest-Chat: Adapting Vision-Language Agents for Interactive Forest Change Analysis}

\author[1]{James Brock\corref{cor1}}
\author[2]{Ce Zhang}
\author[1]{Nantheera Anantrasirichai}
\affiliation[1]{organization={School of Computer Science, University of Bristol},
             addressline={Merchant Venturers Building, 75 Woodland Road},
             city={Bristol},
             postcode={BS8 1UB},
             state={Bristol},
             country={United Kingdom}}

\affiliation[2]{organization={School of Geographical Sciences, University of Bristol},
             addressline={University Road},
             city={Bristol},
             postcode={BS8 1SS},
             state={Bristol},
             country={United Kingdom}}
\cortext[cor1]{Corresponding author. Email address: james.brock@bristol.ac.uk}

\begin{abstract}
The increasing availability of high-resolution satellite imagery, together with advances in deep learning, creates new opportunities for forest monitoring workflows. Two central challenges in this domain are pixel-level change detection and semantic change interpretation, particularly for complex forest dynamics. While large language models (LLMs) are increasingly adopted for data exploration, their integration with vision-language models (VLMs) for remote sensing image change interpretation (RSICI) remains underexplored, especially beyond urban environments. This paper introduces Forest-Chat, an LLM-driven agent for forest change analysis, enabling natural language querying across multiple RSICI tasks, including change detection and captioning, object counting, deforestation characterisation, and change reasoning. Forest-Chat builds upon a multi-level change interpretation (MCI) vision-language backbone with LLM-based orchestration, incorporating zero-shot change detection via AnyChange and multimodal LLM-based zero-shot change captioning and refinement. To support adaptation and evaluation in forest environments, we introduce the Forest-Change dataset, comprising bi-temporal satellite imagery, pixel-level change masks, and semantic change captions generated through human annotation and rule-based methods. Forest-Chat achieves mIoU and BLEU-4 scores of 67.10\% and 40.17\% on Forest-Change, and 88.13\% and 34.41\% on LEVIR-MCI-Trees, a tree-focused subset of LEVIR-MCI. In a zero-shot capacity, it achieves 60.15\% and 34.00\% on Forest-Change, and 47.32\% and 18.23\% on LEVIR-MCI-Trees respectively. Further experiments demonstrate the value of caption refinement for injecting geographic domain knowledge into supervised captions, and the system's limited label domain transfer onto JL1-CD-Trees. These findings demonstrate that interactive, LLM-driven systems can support accessible and interpretable forest change analysis. Datasets and code are publicly available \href{https://github.com/JamesBrockUoB/ForestChat}{here}.

\end{abstract}


\end{frontmatter}


\section{Introduction}
\label{section:intro}
Of the world's terrestrial landmass, forests account for 31 percent of the total area, providing habitat for up to 80\% of land-based species and representing a total carbon stock of 662 gigatonnes in 2020 \citep{canton2021food, lines2022ai}. Forests provide an abundance of resources and ecosystem services including water and nutrient recycling and climate regulation via carbon sequestration, contributing to their central role in adapting to and mitigating climate change \citep{ouaknine2025openforest}. However, anthropogenic pressures and the accelerating frequency of extreme weather events - including industrial logging, agricultural expansion, and wildfires -  are detrimentally impacting the health and integrity of global forest ecosystems \citep{wegler2024potential}.

Monitoring and quantifying the types, extent, and locations of these changes is crucial for both policymakers and researchers \citep{wegler2024potential}. Effective monitoring increasingly relies on automated pipelines as traditional field-surveys become insufficient \citep{ouaknine2025openforest}, with the last decade seeing researchers inundated with remote sensing (RS) imagery capable of providing sub-meter pixel resolution across expansive areas \citep{lines2022ai}. The rich image history, spatial and temporal density, and range of freely available sensor modalities enable more efficient and low-cost monitoring of forest ecosystems \citep{isaienkov2020deep}.

As access to forest-specific datasets and benchmarks has increased, spurred on by tools such as Google Earth Engine (GEE) \citep{gorelick2017google}, a corresponding rise in deep learning (DL) and other artificial intelligence (AI) methods for forest analysis has been observed \citep{isaienkov2020deep}. A review by \cite{yun2024status} highlights a growing publication trend across multiple tasks, including forest resource assessment \citep{nguyen2024multi}, ecological disturbance monitoring \citep{lee2023detection}, parameter retrieval \citep{lin2024model}, and tree organ-level measurement \citep{cai2023tree}. A historical reliance on CNN-based architectures has recently given way to transformers \citep{vaswani2017attention}, attention-based, and hybrid architectures, exemplified by Siamese Attention U-Net models for forest change detection \citep{hewarathna2024change}.

\begin{figure*}
    \centering
    \includegraphics[width=1\textwidth]{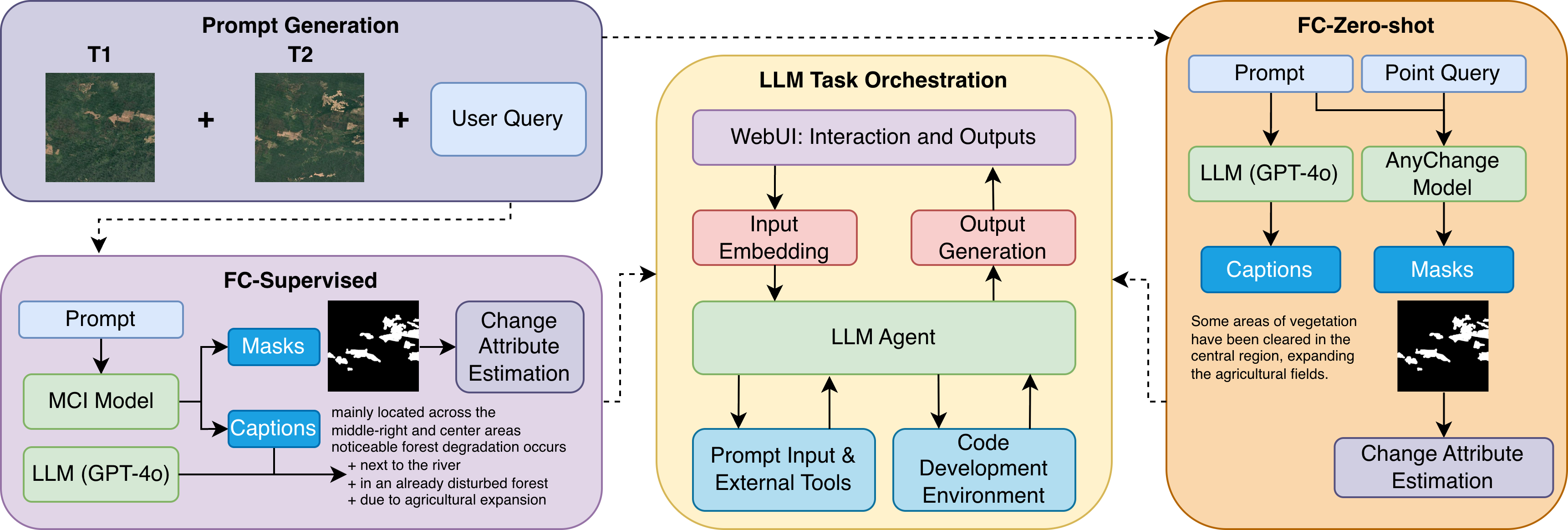}
    \caption[Graphical Abstract]{Overview of the Forest-Chat system. Forest-Chat integrates multiple components: the MCI model \citep{liu2024change} (FC-Supervised) for supervised change captioning, and the combination of AnyChange \citep{zheng2024segment} and GPT-4o \citep{hurst2024gpt}, which together form the FC-Zero-shot module for zero-shot change detection and captioning, as well as optional caption refinement. These components provide the system’s visual perception, while an LLM serves as the reasoning module. The proposed Forest-Change, LEVIR-MCI-Trees, and JL1-CD-Trees datasets are used as a data foundation for training and evaluating Forest-Chat for forest change analysis in a supervised and zero-shot capacity.}
    \label{fig:forest-chat-diagram}
\end{figure*}

The recent trend of exploring large language models (LLMs) as assistants to human domain experts \citep{ge2023openagi} has propagated to remote sensing image change interpretation (RSICI) \citep{liu2024change}. Vision-Language Models (VLMs) have been applied to RS tasks to address longstanding methodological shortcomings in DL-based methods \citep{li2024vision}, particularly the neglect of semantic understanding of objects and their spatial and temporal relationships \citep{li2024prospects}. By harnessing multiple modalities and semantic reasoning, VLMs establish connections between visual concepts and demonstrate improved performance, especially on out-of-distribution data \citep{mu2025globalgeotree}. The culmination of these systems has resulted in chatbots capable of zero-shot and open-vocabulary visual understanding for RS image processing \citep{li2024vision, du2023tree}.

Several review papers have highlighted the opportunities and challenges for large foundation model-assisted RS change detection (RSCD) and RS image change captioning (RSICC) for augmenting domain expert capabilities \citep{zhang2024good, zou2025remote}. Promising directions include leveraging the zero-shot performance of VLMs for RSCD and RSICC \citep{li2024review}, incorporating multimodal data streams \citep{li2024prospects}, and deeper explorations of interactive human-machine collaboration \citep{peng2025deep}. Developing conversational agents with access to RSICI-specific tooling capable of providing interactive and comprehensive change interpretation represents a significant advancement \citep{liu2024change, irvin2025teochat}. Models capable of dual change detection and captioning (CDC) that incorporate LLM agents for task orchestration are of particular interest here - Change-Agent \citep{liu2024change} specifically represents a paradigm shift, being the first VLM capable of dual CDC embedded within an agent for spatio-temporal RS tasks.

While frameworks like Change-Agent demonstrate the potential of natural language interfaces for RS tasks, they often require large task-specific training datasets to align change detection outputs with semantic descriptions \citep{irvin2025teochat}. Recent multimodal LLMs (MLLMs) capable of strong single-image RS captioning without task-specific training \citep{ibrahimcomparative, zhang2024good} offer a promising alternative, naturally motivating their application in bi-temporal change settings. However, zero-shot MLLM-based RS change captioning remains largely underexplored - where benchmarked using general prompting strategies, even large open-source MLLMs fail to capture meaningful semantic differences between bi-temporal images, producing generic or spatially inconsistent descriptions \citep{leon2026describing, soni2025earthdial}. This is attributed to a lack of domain-specific grounding, suggesting that further adaptation - whether through fine-tuning or domain-informed prompting - is needed to unlock the potential of MLLMs for zero-shot change captioning.

In contrast, recent work has demonstrated stronger potential for pixel-level change detection. AnyChange \citep{zheng2024segment} introduces a promptable, zero-shot semantic change detection model extending the Segment Anything Model (SAM) \citep{kirillov2023segment} by employing a training-free bi-temporal latent matching method. Although such methods offer training-free zero-shot capabilities, effective domain-specific change interpretation - particularly in forest applications - remains challenging due to subtle structural variations, seasonal phenology, and heterogeneous disturbance patterns, and thus still benefits greatly from purpose-built architectures and tailored datasets.

To address these challenges, this work introduces Forest-Chat (Fig. \ref{fig:forest-chat-diagram}), a Change-Agent inspired VLM-agent system capable of both supervised and zero-shot interactive forest change analysis. The Multi-level Change Interpretation (MCI) model is retained as the `eyes' of the system for dual supervised CDC, while an LLM serves as the `brain' to orchestrate reasoning and dialogue. A zero-shot perception pipeline (FC-Zero-shot) is additionally incorporated, combining AnyChange for zero-shot and interactive point-query based change segmentation with GPT-4o for change captioning and refinement. To the best of our knowledge, Forest-Chat is the first LLM-driven agent that simultaneously integrates both supervised and zero-shot visual perception modules for RSICI. The system is augmented with analytical tools that process generated change masks to support deeper interpretation of forest change dynamics.

The paper introduces the Forest-Change dataset, the first dataset specifically designed for forest change analysis providing both bi-temporal pixel-level change masks and semantic change captions. Forest-Chat offers a conversational interface for interactive exploration of forest change events, reducing cognitive load and manual analysis effort, opening opportunities for expert-in-the-loop analysis of long-term and heterogeneous forest dynamics. 

Extensive experiments evaluate Forest-Chat’s supervised and zero-shot capabilities on the Forest-Change and LEVIR-MCI-Trees datasets. Zero-shot MLLM change captioning is further benchmarked under general and style-guided prompting strategies, with the effect of domain-specific caption refinement additionally analysed. Additional studies investigate backbone model size, multi-task loss balancing and gradient conflict resolution strategies, and cross-domain transfer robustness using the JL1-CD-Trees dataset. Forest-Chat's utility as an interactive tool for forest change analysis is further demonstrated through qualitative case studies.

The contributions of this work are outlined below:
 
\begin{itemize}
    \item \textbf{Forest-Chat} is presented as an interactive forest change analysis system integrating multiple visual perception modules, and as an empirical exploration of adapting VLM-based RS agents to forest change analysis. This is a non-trivial task given that annotated forest change data is scarce, and the ecological diversity of forest change processes — spanning abrupt clearance events to gradual degradation and reforestation — is poorly represented in existing RSICI datasets and systems. FC-Supervised provides supervised pixel-level CDC via the MCI model, while FC-Zero-shot combines AnyChange for pixel-level change detection with GPT-4o for change captioning and refinement. An LLM orchestrates reasoning and dialogue, enabling flexible exploration of temporal RS imagery. Extensive experiments across three datasets evaluate architectural components, backbone scaling, multi-task learning strategies, and cross-domain generalisation.
    \item \textbf{Zero-shot MLLM change captioning} is empirically analysed as an underexplored task, benchmarking an MLLM against supervised captioning under general and style-guided prompting strategies, with the effect of domain-specific caption refinement additionally analysed.
    \item \textbf{Forest-Change} is proposed as the first dedicated dataset for RSICI in forest applications, incorporating bi-temporal image pairs with corresponding change masks and captions describing the degree, location, and patterns of observed deforestation. 
    \item \textbf{LEVIR-MCI-Trees} is curated from LEVIR-MCI by retaining samples where tree changes are semantically present, yielding a benchmark that pairs urban-focused change detection masks with tree-aware captions. It provides a large-scale complement to Forest-Change for evaluating captioning generalisation beyond purely forest scenes.
    \item \textbf{JL1-CD-Trees} is introduced as a woodland-focused subset of JL1-CD providing a rigorous out-of-distribution benchmark for cross-domain evaluation, assessing model robustness to distribution shift from sensor differences, seasonal variability, and atmospheric artifacts.
\end{itemize}

\section{Related Work}
\label{section:related_work} 

This work concentrates on vision-language agents (VLAs) capable of multi-task learning for remote sensing image change interpretation (RSICI) tasks. Research directions into RS-VLAs are increasing in both maturity and diversity, but the areas relevant to this work include RSICI, foundation and vision-language models (VLMs), and large language model (LLMs) agents. 

\subsection{Remote Sensing Image Change Interpretation (RSICI)}
\label{section:background-rmici}
RSICI commonly encompasses the tasks of RS change detection and change captioning (CDC), applicable across diverse remote sensing data modalities and resolutions \citep{liu2024change}. RSCD is a fundamental spatio-temporal task aimed at identifying changes between multi-temporal images, either as binary predictions or across multiple categories  \citep{liu2025remote, li2025cd4c}. Deep learning approaches have progressed from RNN- and CNN-based models to transformers incorporating self-attention, enabling the capture of global context and long-range dependencies crucial for complex scene interpretation \citep{tao2025advancements}. Progress has been further accelerated by multi-scale feature fusion, Siamese networks, and pretrained backbone networks \citep{shi2024multi}. Zero-shot RSCD was recently achieved by the \textit{AnyChange} model \citep{zheng2024segment}, which leverages SAM \citep{kirillov2023segment} for annotation-free change analysis. 

RSCD is fundamentally limited to explaining \textit{what} has changed, rather than \textit{why} and \textit{how}. RSICC addressed this by combining image captioning with change detection, demanding precise visual change recognition and robust language generation \citep{liu2025remote, tao2025advancements}. Contemporary methods typically adopt a three-stage architecture, incorporating CNNs, transformers, and Vision Transformers (ViTs) to extract semantic change representations and produce coherent descriptions -  a paradigm established by RSICCformer \citep{liu2022remote} and its accompanying LEVIR-CC benchmark. Subsequent work refined this paradigm - CD4C \citep{li2025cd4c} integrated a supervised change-detection branch to handle samples with varying change reliability, and KCFI \citep{yang2025enhancing} added pixel-level change detection for improved semantic grounding - while MCI \citep{liu2024change} was the first framework to fully unify RSCD and RSICC in a multi-task learning setting, simultaneously generating change masks and textual descriptions, supported by the LEVIR-MCI dataset. Unifying these tasks introduces the challenge of balanced optimisation, as RSCD and RSICC compete for shared representations - requiring careful consideration of loss balancing and gradient conflict resolution strategies \citep{shi2024multi, crawshaw2020multi}.

\subsection{Vision-Language Models for Remote Sensing}
\label{section:background-vlms}

Vision–language models (VLMs) combine instruction-following with visual feature analysis, typically pairing a pretrained visual encoder - such as CLIP model \citep{radford2021learning}, or a ViT - with an LLM (e.g. GPT \citep{openai2023}, LLaMa \citep{touvron2023llama}) through a projection or transformation module that maps visual embeddings into the language model's input space \citep{li2024unirs, tao2025advancements}. Recent work has explored improved vision-language pretraining strategies for aligning visual and textual representations \citep{qin2025fine}. This alignment enables generalised tasks such as visual question answering, captioning, and zero-shot transfer learning \citep{lu2025vision, li2024review}. These capabilities have made VLMs an increasingly common choice for RS applications \citep{zhang2025georsmllm, liu2024rsunivlm, li2024unirs, hu2025rsgpt, zhang2024earthgpt}, which benefit from domain-specific pretraining across task-agnostic objectives, varying sensors, resolutions, and imaging geometries, including multimodal fusion of complementary data sources \citep{yang2024modality, lu2025vision}. Self-supervised and zero-shot learning further reduces reliance on exhaustive annotation \citep{peng2025deep, lu2025vision}, while text prompting and question-answering interactions support context-aware interpretation, helping models link ambiguous visual patterns to high-level semantic concepts such as deforestation \citep{li2024prospects}.

Despite strong performance on natural imagery, general-purpose multimodal LLMs (MLLMs) such as GPT-4V and LLaVa \citep{openai2023, bazi2024rs} underperform on several RS and change-related tasks due to domain mismatch \citep{zhang2024good}. One response has been to adapt LLM-driven, task-specific spatio-temporal VLMs to RS vision-language understanding tasks \citep{liu2025remote}, including models such as KCFI \citep{yang2025enhancing} and Semantic-CC \citep{zhu2024semantic}, that fine-tune visual modules or lightweight LLM components for change-focused tasks. Related efforts such as Change-UP \citep{yang2025change} and ChangeMinds \citep{wang2024changeminds} target joint RSCDC within unified, end-to-end architectures. However, these approaches remain narrow in objective scope, relying on specialised labelled datasets that limit generalisation (e.g. RSGPT \citep{hu2025rsgpt}) \citep{liu2025remote}. A broader direction aims to unify RS tasks within vision-language foundation models \citep{li2024unirs}. GeoChat \citep{kuckreja2024geochat} and EarthGPT \citep{zhang2024earthgpt} adapt LLaVa-style conversational VLMs for multi-task, grounded region-aware dialogue and multi-sensor interpretation while RSUniVLM \citep{liu2024rsunivlm} and UniRS \citep{li2024unirs} extend this to pixel-level understanding and spatio-temporal multi-image analysis respectively.

Most recently, LLM-driven RS vision–language agents extend these models by integrating perception, reasoning, and tool use into multistep analytical workflows, supporting tasks such as temporal scene understanding and change detection through multi-turn conversations \citep{irvin2025teochat, liu2024change}. Tree-GPT \citep{du2023tree} is one such example, combining LLM orchestration with knowledge bases and vision modules for interactive forest scene analysis, albeit lacking a temporal component.

\subsection{LLM Agents for Remote Sensing}
\label{section:background-llm-agents}

AI agents link perception and action through iterative reasoning to achieve user-defined goals \citep{xu2024rs}. Utilising LLMs pretrained on massive text corpora, they can decompose tasks and interact dynamically across steps with external data, models, or tools \citep{liu2025remote}, serving as modular orchestrators for complex workflow execution \citep{liu2024change, schick2023toolformer, yao2023react, guo2024remote}.

Within RS, LLM-based agents are still in their infancy \citep{xu2024rs}. They are typically task-specific, relying on strict prompt templates and few-shot examples to guide reasoning \citep{guo2024remote}. Agents execute tasks by selecting from specialised visual toolkits, composable into sequential pipelines, as in Change-Agent \citep{liu2024change} and RS-Agent \citep{xu2024rs} to accomplish multiple objectives. This modular design - often augmented with knowledge bases - reduces hallucinations and enables flexible outputs \citep{guo2024remote, liu2025remote}. RS-specific LLM agents generally fall into two broad paradigms: modular tool-using systems that rely on external LLMs to orchestrate specialised vision modules; and end-to-end instruction-tuned VLMs that embed temporal or multimodal reasoning directly within the model \citep{liu2024change, xu2024rs, deng2025changechat, irvin2025teochat}.

Key limitations span both paradigms - modular systems face restricted end-to-end flexibility \citep{deng2025changechat}, while flexible task orchestration, mitigating task decomposition failures, and robust tool selection remain challenging across approaches \citep{xu2024rs}. TEOChat \citep{irvin2025teochat} bridges these design philosophies through instruction tuning over variable-length temporal image sequences, enabling stepwise reasoning across multi-date imagery while maintaining modular interpretability. RS-Agent \citep{xu2024rs} addresses these issues by using an LLM controller that selects from a suite of geospatial tools and composes multistep workflows, supported by a retrieval-augmented knowledge graph for domain-specific reasoning. Change-Agent \citep{liu2024change} is designed specifically for bi-temporal change analysis, integrating an LLM with change-focused reasoning modules to support interactive and interpretable assessments of temporal differences in RS imagery.

Comparing these agents is challenging due to differences in task scope, input modalities, and evaluation setups, making it difficult to assess performance and agentic reasoning across approaches  \citep{shabbir2025thinkgeo}. To date, only ChangeMinds \citep{wang2024changeminds} and Change-UP \citep{yang2025change} benchmark against Change-Agent's MCI model \citep{liu2024change} for joint RSCD and RSICC, with both reporting modest gains. However, neither incorporates an LLM-based agent, or is publicly available. Other frameworks either lack temporal reasoning \citep{guo2024remote, du2023tree}, or offer limited CDC capabilities \citep{irvin2025teochat, deng2025changechat}, leaving Change-Agent as the most appropriate baseline identified.

\section{Data}
\label{section:data}

\subsection{LEVIR-MCI-Trees}
\label{section:data-levir-mci-trees}

When developing the robust MCI model at the heart of the Change-Agent system, \cite{liu2024change} proposed the LEVIR-MCI dataset, which comprises both pixel-level change information contained within change detection masks and descriptive semantic-level captions. Each of the 10,077 examples contains bi-temporal image pairs with a variable time span of 5-15 years, a corresponding annotated mask, and five annotated captions. Each image spans 256$\times$256 pixels at a high spatial resolution of 0.5m/pixel. LEVIR-MCI extends LEVIR-CC \citep{liu2022remote} by providing each pair of bi-temporal images with change detection masks that explicitly highlight changes to roads and buildings, with over 40,000 changed roads and buildings annotated. The five captions for each image pair are intended to provide diverse annotations from varying perspectives to bolster its utility for change detection. 

For the purposes of this study, a subset of this dataset was taken that discards image pairs that contain no changes to tree cover, hereby referred to as \textit{LEVIR-MCI-Trees}. Filtering is performed based on the contents of the captions for each example. If none of the captions for an example contains one of: `\emph{tree, trees, wood, woods, woodland, wooded, forest, forests, jungle, jungles}` then it is removed. Post filtering, the resulting  subset contains 2,305 examples distributed across training, validation and test sets comprising 66\%, 16\%, and 18\% (1518, 374, 413) of the dataset respectively. Due to the dataset only containing pixel-level annotations for roads and buildings, it is unsuitable for benchmarking Forest-Chat's performance for deforestation segmentation directly, but can be used to gauge the architecture's generalisation capabilities to new geographical domains. However, this subset will be suitable for comparing the performance of tree disturbance captioning and highlighting any differences in the difficulty of pixel-level annotation between urban and forest contexts. Segmentation performance in urban settings is generally higher than in natural landscapes due to simpler object geometries and clearer boundaries \citep{wang2025multi}, leading to a preference in benchmarking model performance within urban contexts.

\subsection{Forest-Change}
\label{section:data-forest-change}
This section outlines the work undertaken to create a novel change caption and detection dataset for forest cover analysis, highlighting the creation process, along with potential limitations.

\subsubsection{Data Curation}
\label{section:data-curation}

To the best of our knowledge, no existing benchmark dataset explicitly supports joint forest change detection and captioning (CDC). \cite{lines2022ai} highlight the broader scarcity of RS forest monitoring benchmark datasets, and while recent initiatives such as OpenForest \citep{ouaknine2025openforest} catalogue available resources, few datasets target forest change specifically.

Among existing datasets, CAM-ForestNet \citep{debus2024labelled, hartanti2023multimodal} is intended to support deforestation segmentation using time-series imagery, providing both change masks and deforestation driver labels. These works largely gather imagery from Google Earth Engine (GEE) and then verify forest cover changes through the open-source Global Forest Watch (GFW) platform \citep{gfw2014}, which serves as a crucial tool for tracking global forest health. However, it is not designed for joint change detection and captioning tasks, and issues such as misaligned polygons, corrupted or inconsistent imagery, and variable temporal sequences limit its suitability for RSICI model development.

\cite{hewarathna2024change} introduce a dataset focused solely on forest cover change detection, albeit without deforestation driver labels that could support caption generation. It comprises variable-length temporal RGB stacks of tropical and subtropical forest loss from GEE \citep{gorelick2017google}, with a temporal resolution of one year across multiple locations, informed by the WWF Deforestation Fronts report \citep{taylor2015deforestation}. The authors crop, augment, and segment the original images into 480$\times$480 patches at a medium spatial resolution of $\sim$30m/pixel, resulting in 334 annotated bi-temporal image pairs with binary change masks from an original set of 1500. 

\begin{figure}
    \centering
    \includegraphics[width=1\linewidth]{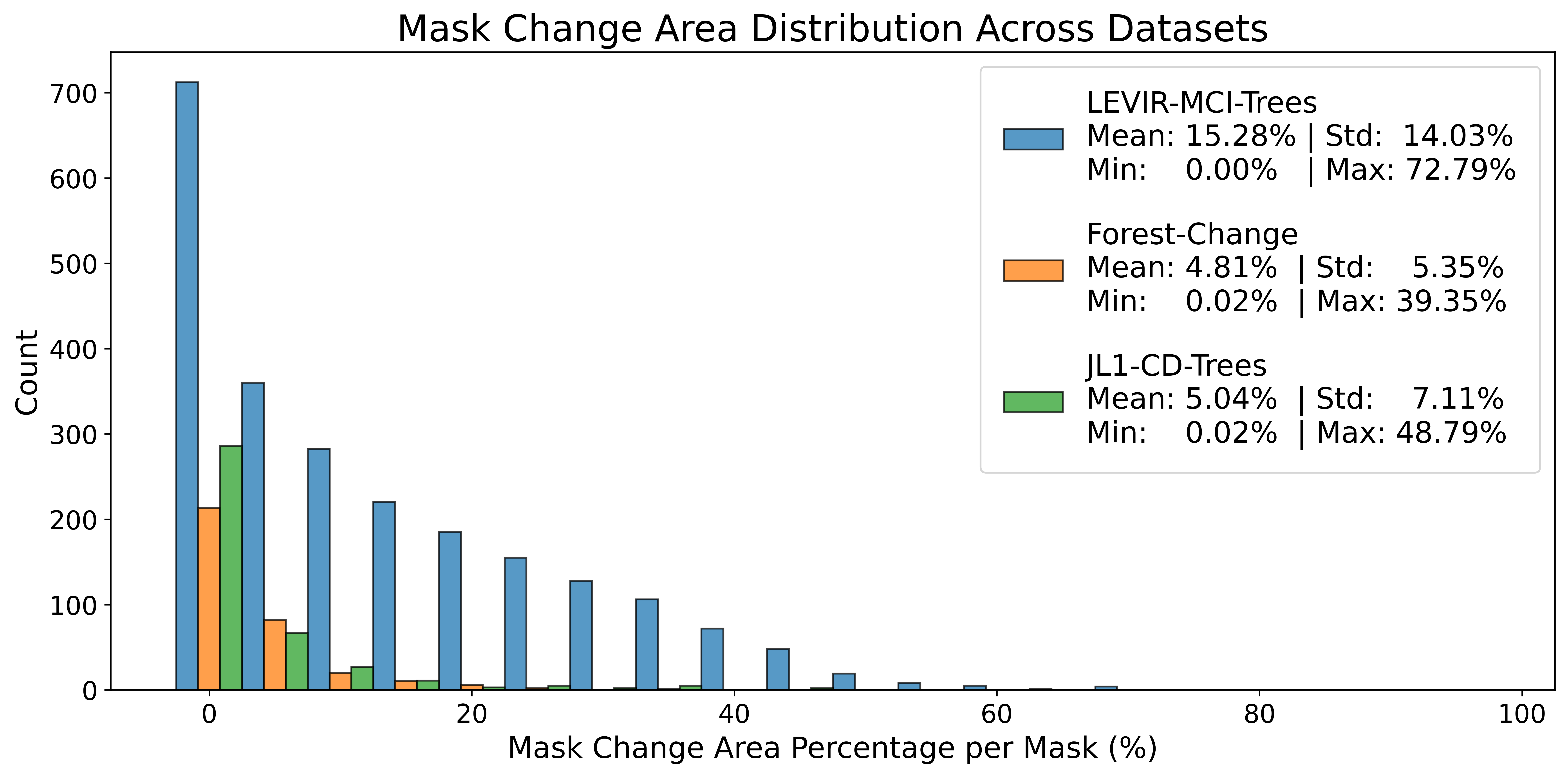}
    \caption[Dataset Segmentation Mask Statistics]{Summary statistics of change cover in segmentation masks for the Forest-Change, LEVIR-MCI-Trees, and JL1-CD-Trees datasets.}
    \label{fig:dataset_mask_distribution}
\end{figure}

\subsubsection{Preprocessing}
\label{section:preprocessing}

Due to overlapping crops and the conversion of temporal image sequences to bi-temporal image pairs, both scene diversity and change mask geometry are limited. As shown in Figure \ref{fig:dataset_mask_distribution}, most masks in the Forest-Change dataset contain less than 5\% new deforestation, with a maximum of ~40\%, resulting in a heavily imbalanced dataset that necessitates per-class evaluation rather than global pixel accuracy. In contrast, the LEVIR-MCI-Trees dataset exhibits a higher and more consistent foreground proportion (mean 15.28\%, max 72.79\%) and more regular object geometries, making change segmentation conceptually easier.

The Forest-Change dataset is organised to mirror the directory structure of LEVIR-MCI-Trees to maintain consistent data loading. Three splits were created for training, validation, and testing at an approximate ratio of 80\%:10\%:10\% (270, 31, and 33 samples, respectively). All images are resized to 256$\times$256 pixels to match LEVIR-MCI-Trees, and pixel-wise change masks are binarised to indicate change (1) or no-change (0). The bi-temporal image pairs are pre-aligned, and no additional geometric registration or atmospheric correction was applied. Images were normalised using dataset-specific per-channel mean and standard deviation statistics. No additional radiometric correction or cloud masking was performed, although a few samples exhibit partial cloud occlusion.

\subsubsection{Caption Generation}
\label{section:data-caption-generation}

\begin{figure}
    \centering
    \includegraphics[width=1\linewidth]{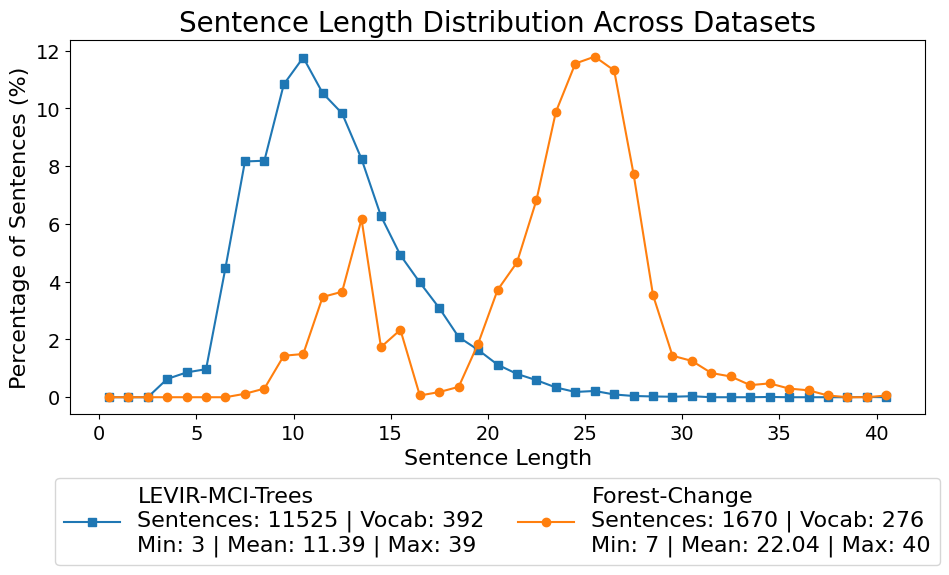}
    \caption[Dataset Sentence Statistics]{Summary statistics of  captions for Forest-Change and LEVIR-MCI-Trees datasets.}
    \label{fig:dataset_captions_distribution}
\end{figure}

As shown in Figure \ref{fig:dataset_captions_distribution}, captions for the Forest-Change dataset are skewed toward descriptions of minor forest loss and recurring loss patterns. This reflects the limited scene diversity introduced by the cropping-augmentation strategy adopted by \cite{hewarathna2024change}, as discussed previously in Section \ref{section:data-curation}. Captioning large volumes of remote sensing change imagery is labour-intensive for human annotators, especially when scene diversity is limited. Therefore, rule-based caption generation is proposed as a way to ensure minimal requirements for semantic context and sentence composition are met. Human annotations remain valuable for improving linguistic and semantic variation, and providing geographic context.

Caption generation was conducted using a two-stage approach combining manual annotation with rule-based caption synthesis. A custom Streamlit-based application was developed to facilitate captioning due to its suitability for rapid interactive prototyping and its prior integration within the Change-Agent framework \citep{liu2024change}.

The application prompts the user to select a dataset directory, following the LEVIR-MCI-Trees' folder structure (also used for Forest-Change). Previously labelled samples can be skipped to support incremental annotation. The annotator is iteratively presented with bi-temporal image pairs and the corresponding change mask, and asked to provide a single short-to-moderate length caption describing the observed change. Upon submission, four additional captions are automatically generated based on properties extracted from the change mask - producing a total of five captions. 

For automated caption generation, the percentage of newly deforested area, the size of individual change patches, and their spatial distribution are computed from the mask. The percentage of forest loss is binned into adjective-based descriptors aligned with the model vocabulary. A rule-based caption generator was designed to reduce labelling burden, and enforce sufficient change descriptions are generated, producing complementary captions to the human-provided ones. Although LLMs are increasingly used for synthetic text generation in NLP tasks \cite{yu2023large}, they were not adopted here due to the risk of introducing prompt-induced bias and the limited scope of the project in which to mitigate it.

\begin{figure}[ht!]
  \centering
  \begin{subfigure}[t]{0.45\textwidth}
    \centering
    \caption{three noted areas of loss in the top left top right and centre on the edge of previous clearings new loss is moderate whilst overall loss is considerable}
    \includegraphics[width=\linewidth]{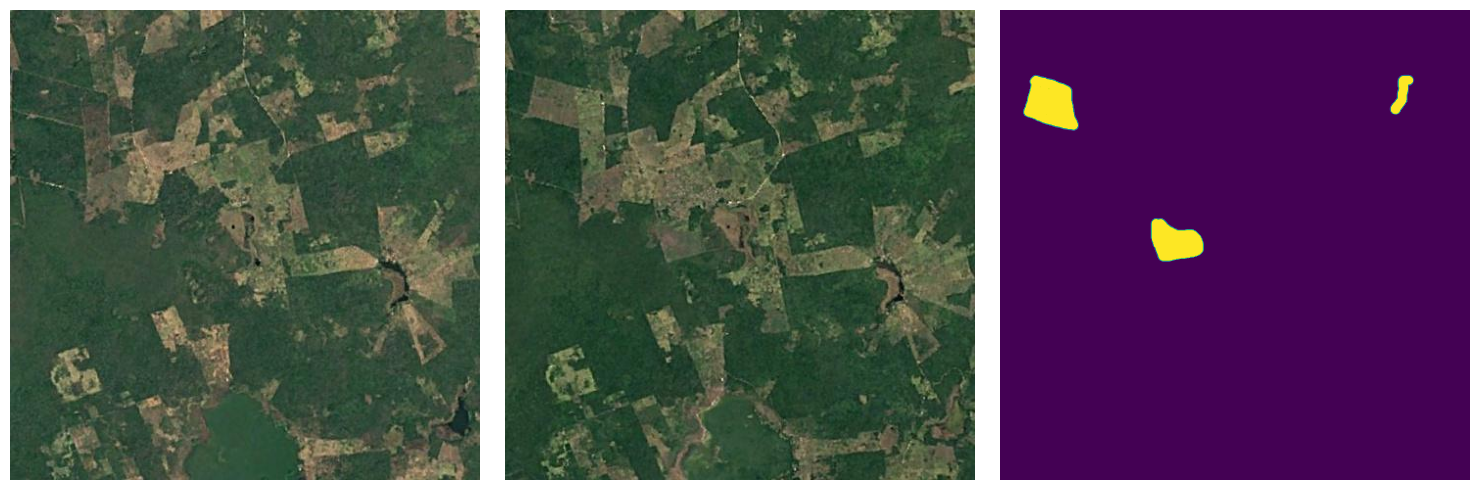}
  \end{subfigure}
  \hspace{0.05\textwidth}
  \begin{subfigure}[t]{0.45\textwidth}
    \centering
    \caption{slight forest degradation is noted scattered across multiple regions occurring in many small patches which are highly varied in size}
    \includegraphics[width=\linewidth]{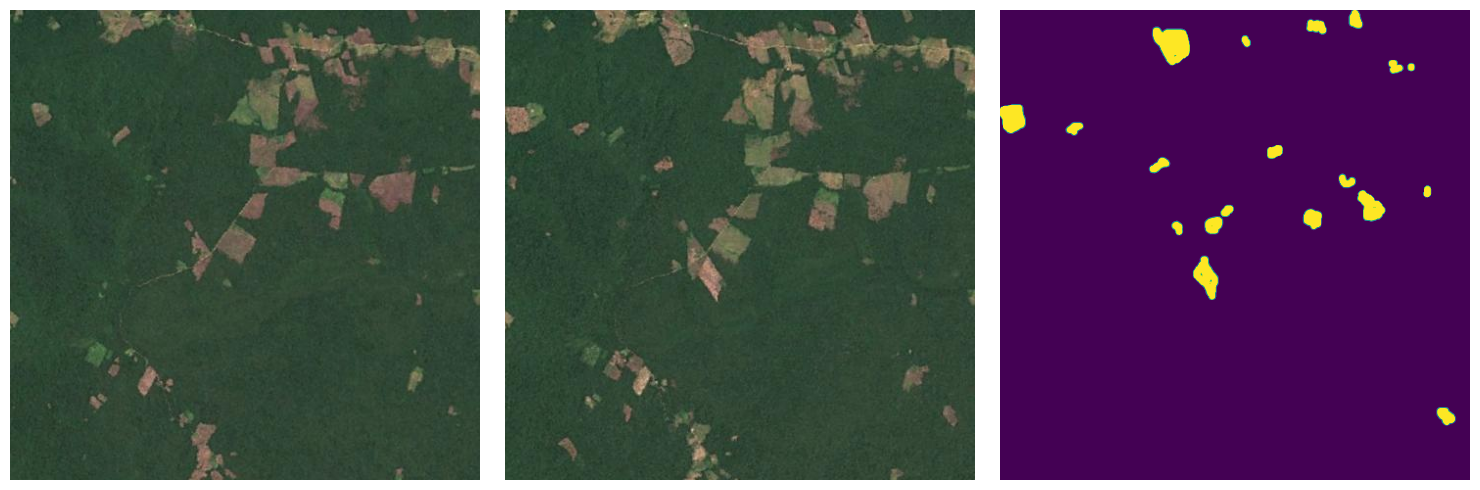}
  \end{subfigure}

  \begin{subfigure}[t]{0.45\textwidth}
    \centering
    \caption{some modest forest loss is detected largely concentrated in the top-left and top-center sections occurring in many small patches which are displaying large variations in size}
    \includegraphics[width=\linewidth]{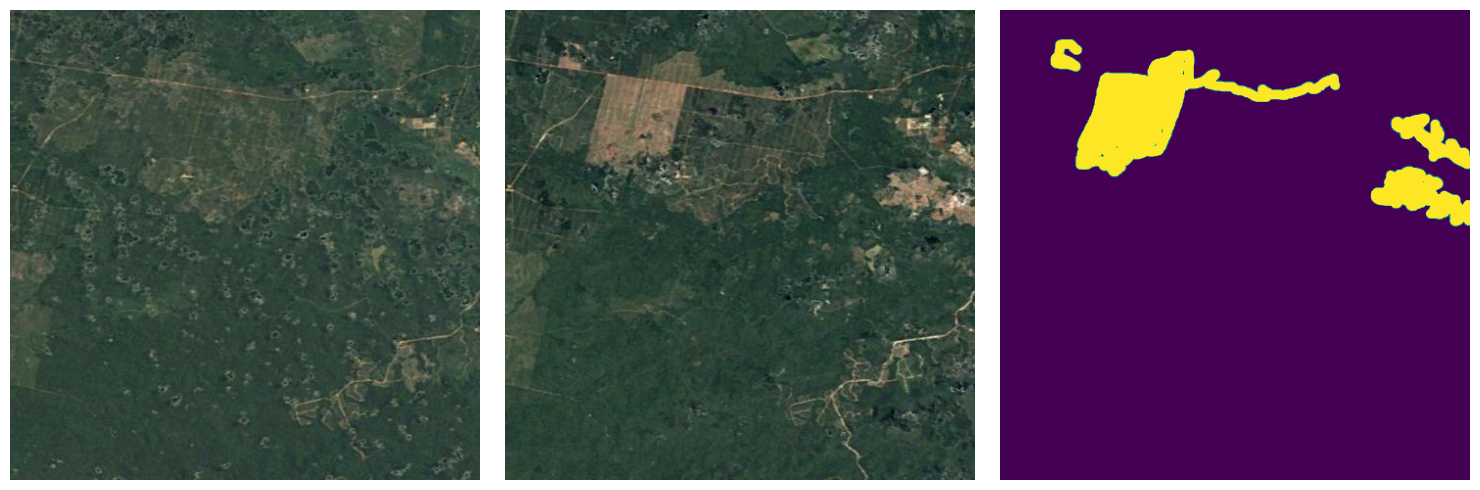}
  \end{subfigure}
  \hspace{0.05\textwidth}
  \begin{subfigure}[t]{0.45\textwidth}
    \centering
    \caption{minor forest loss is visible occurring in many small patches which are showing some variation in size mainly located in the top-left area}
    \includegraphics[width=\linewidth]{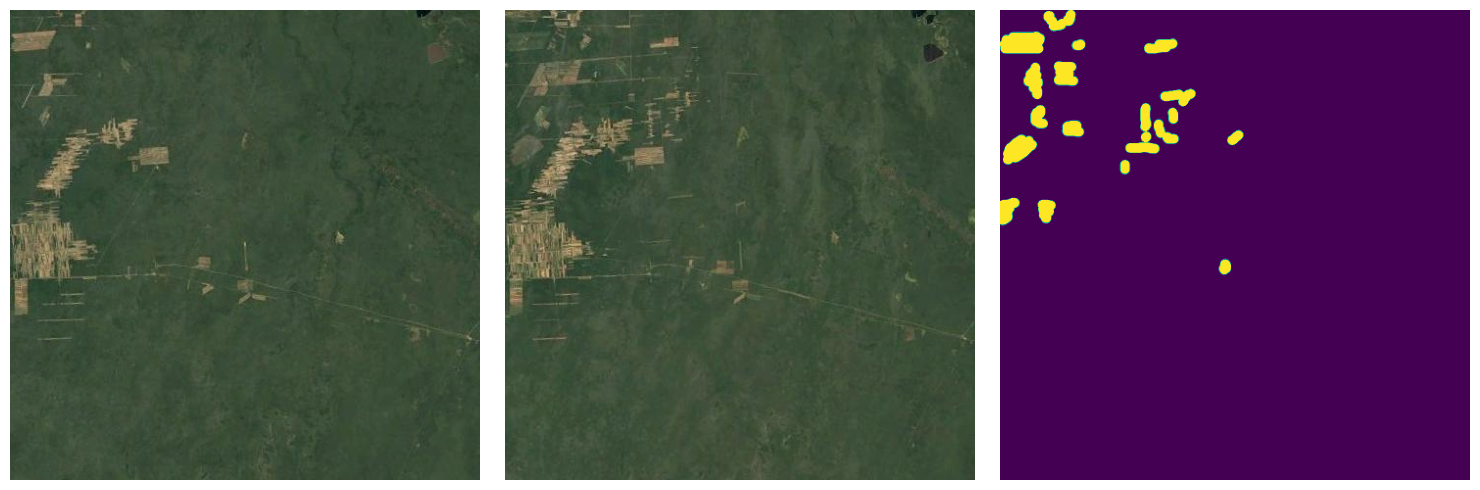}
  \end{subfigure}
  \caption[Forest-Change Dataset Examples]{Four examples from the Forest-Change dataset, that each display a randomly selected caption, combined with image A, image B, and the corresponding forest change mask.}
  \label{fig:combined_examples}
\end{figure}

Figure \ref{fig:dataset_captions_distribution} compares caption characteristics between Forest-Change and LEVIR-MCI-Trees. Captions in LEVIR-MCI-Trees are generally shorter and exhibit greater lexical diversity, while Forest-Change captions display a bimodal length distribution that reflects the combination of rule-based and human-generated annotation. By combining the unique characteristics of the datasets, evaluating model performances becomes more robust. Figure \ref{fig:combined_examples} presents representative caption examples from the Forest-Change dataset, illustrating the range of forest loss descriptions.

\subsection{JL1-CD-Trees}
Remote sensing datasets often contain varying degrees of noise, seasonal variability, cross-sensor inconsistencies, and domain shifts. To evaluate Forest-Chat's robustness and domain transfer capabilities for unseen forest change detection scenarios, we introduce JL1-CD-Trees. The JL1-CD dataset was proposed by \cite{liu2025jl1} as a benchmark for evaluating CD algorithms. It comprises 5,000 satellite image pairs captured by the Jilin-1 satellite, across various regions of China between early 2022 and the end of 2023. The imagery has sub-meter spatial resolution ranging from 0.5 to 0.75 meters. The dataset includes a diverse set of anthropogenic and natural surface features, including buildings, roads, hardened surfaces, woodlands, grasslands, croplands, water bodies, and photovoltaic (PV) panels. Each change mask may contain multiple surface change types.

To create JL1-CD-Trees, the dataset was manually filtered to only retain samples containing changes between woodland and other natural surface types (e.g. grassland), thereby maximising domain alignment with Forest-Change. The resulting subset presents several challenges, including seasonal variation, atmospheric artifacts (e.g. cloud cover and haze), and both gains and losses in tree cover. After filtering, the dataset contains 408 samples, split into training, validation, and test sets with an approximate 60:20:20 ratio (244, 81, and 83 samples, respectively). To ensure consistency with Forest-Change, identical preprocessing steps are applied, including image rescaling, change mask binarisation, and pixel value normalisation. Figure \ref{fig:dataset_mask_distribution} highlights that JL1-CD-Trees and Forest-Change have similar data distributions and class imbalances. JL1-CD-Trees is intended solely for evaluating domain transfer in forest change detection; therefore no captions are provided.

\section{Methodology}
\label{section:methodology}

This section outlines the components of the Forest-Chat agent. The goal of the Forest-Chat agent is to provide a conversational interface for bi-temporal RS forest change analysis through joint pixel-level change detection and change captioning (CDC). Building on the Change-Agent framework \citep{liu2024change}, Forest-Chat integrates a domain-specific VLM (the MCI model) for supervised CDC, a zero-shot foundation model (AnyChange) for class-agnostic pixel-level change localisation, and GPT-4o for zero-shot change captioning and refinement. These supervised and zero-shot configurations are referred to as FC-Supervised and FC-Zero-shot respectively. To adapt Forest-Chat for forest change analysis, the MCI model is trained on the Forest-Change and LEVIR-MCI-Trees datasets to produce structured change masks and semantic change descriptions. An interactive interface is additionally developed for user-provided point prompts to guide object-oriented change detection with AnyChange. An LLM controller oversees instruction interpretation, tool selection, and analytical reasoning across all modules. The system is further augmented with forest-specific analytical functions, including deforestation area estimation and patch-level statistics, enabling both low-level perception and high-level reasoning within a unified agentic framework.

\begin{figure*}[ht!]
    \centering
    \begin{subfigure}[b]{0.9\textwidth}
        \centering
        \includegraphics[width=1\textwidth]{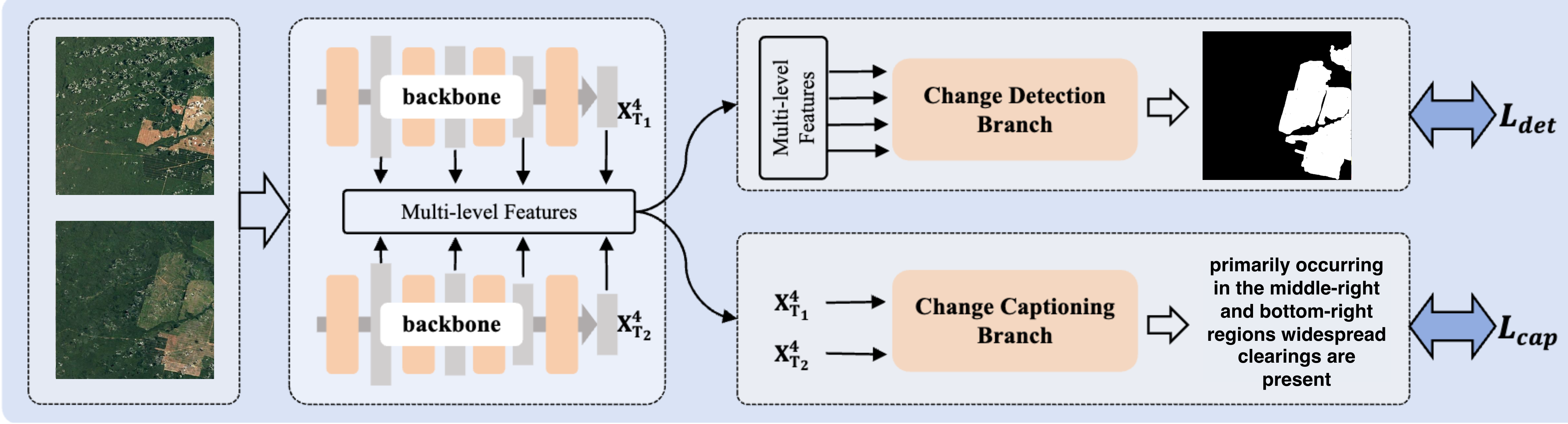}
        \caption{MCI model overview. Adapted from ChangeAgent \citep{liu2024change}.}
        \label{fig:mci-forestchat}
    \end{subfigure}

    \centering
    \tikz{\draw[dashed, line width=0.5pt] (0,0) -- (0.8\textwidth,0);}
    \vspace{0.3cm}

    \begin{subfigure}[b]{1\textwidth}
        \centering
        \includegraphics[width=0.7\textheight]{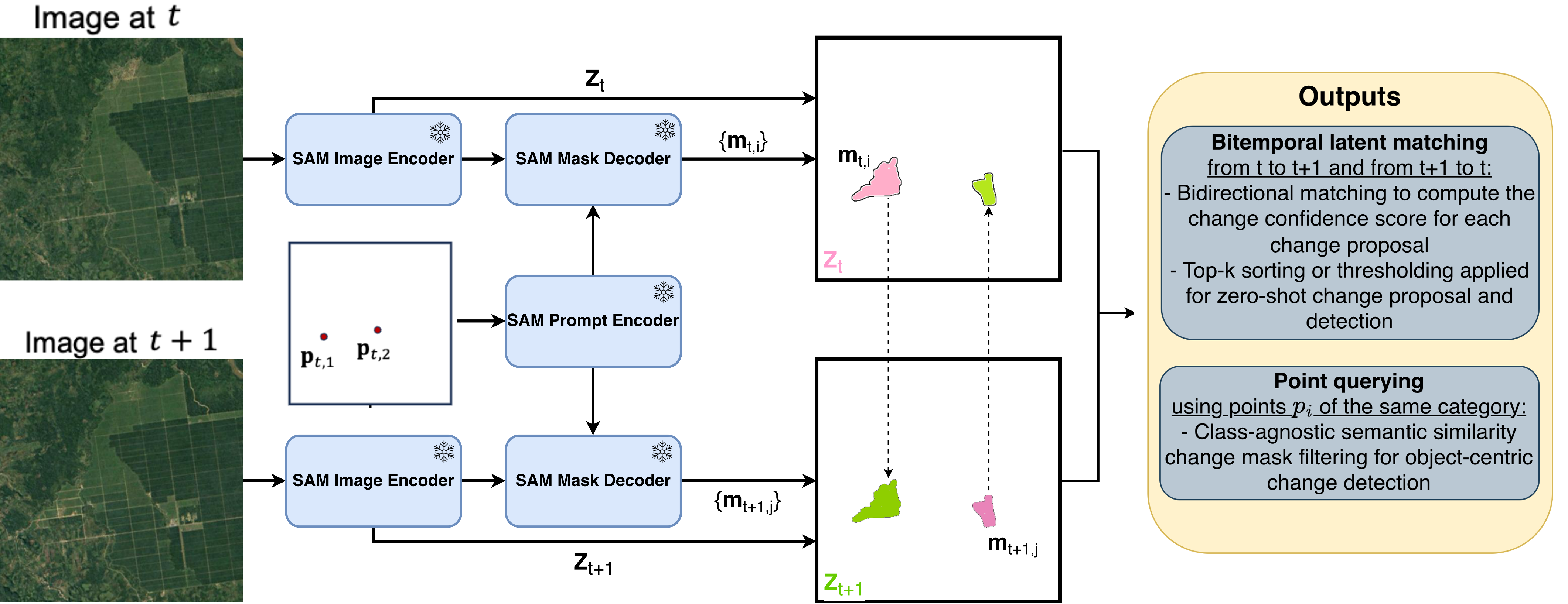}
        \caption{AnyChange overview.}
        \label{fig:any-change-diagram}
    \end{subfigure}
    \caption[Forest-Chat Visual Perception Modules]{Overview of the two visual perception models available to the Forest-Chat agent. (a) Highlights the overall structure of the MCI model (b) Introduces a simplified overview of the AnyChange model. GPT-4o is additionally integrated for zero-shot captioning and captioning refinement, but a diagram is not provided due to its proprietary status.}
    \label{fig:perception-modules-diagram}
\end{figure*}

The overall system diagram is presented in Figure \ref{fig:forest-chat-diagram}, with Figure \ref{fig:perception-modules-diagram} providing a simplified overview of the incorporated perception models.

\subsection{Multi-level Change Interpretation Model}
\label{section:methodology-mci-model}
The MCI model is a dual-branch VLM designed to process pairs of bi-temporal RS images, as shown in Figure \ref{fig:mci-forestchat}. A Siamese SegFormer B1 backbone \citep{xie2021segformer} extracts multi-scale semantic features from each image, providing both low-level detail and high-level contextual information, forming the shared representation from which two task-specific branches are derived.

The central innovation is the \textbf{Bi-temporal Iterative Interaction (BI3) layer}, embedded within both branches to enhance and fuse bi-temporal features through \textit{Local Perception Enhancement} (LPE), and \textit{Global Difference Fusion Attention} (GFDA) to extract discriminative features of interest. LPE applies multi-scale convolutions to capture local change variations, while GDFA uses feature differencing and attention to highlight regions of true change while suppressing noise. Three BI3 layers are stacked to progressively strengthen feature discrimination across both branches.

The \textbf{Change detection branch} (top right, Fig. \ref{fig:mci-forestchat}) applies stacked BI3 layers with residual connections, before employing Convolution-based Bi-temporal Fusion (CBF) modules across four spatial scales and a deconvolution pathway to produce precise, high-resolution change masks. The \textbf{Change captioning branch} (bottom right, Fig. \ref{fig:mci-forestchat}) refines semantic features through stacked BI3 layers, projects them into the textual domain via a convolution-based projection layer, and decodes them using a transformer decoder to generate natural language descriptions of observed changes.

\subsection{LLM Agent Task Orchestration}
\label{section:methodology-llm-orchestration}

The LLM functions as the controller of Forest-Chat (Fig. \ref{fig:forest-chat-diagram} right side), interpreting user instructions, selecting appropriate perception modules, directing task execution, and synthesising results into natural language responses. While some recent LLMs offer native visual perception \citep{openai2023, hurst2024gpt}, these capabilities are not yet optimally suited to RS imagery \citep{zhang2024good}, motivating the integration of specialised external tools including the MCI model, AnyChange, GPT-4o, and forest-specific analytical functions. By generating and executing Python programs, the LLM can invoke these tools as needed, enabling automated multistep workflows without human intervention. System prompts with few-shot examples, retry mechanisms, and output validation are provided to support forest change analysis workflows, ensuring reliable tool invocation and consistent output formatting.

\textit{ChatGPT-4o-mini} serves as the LLM backbone, selected to balance performance, speed, and cost. GPT-4 models have demonstrated strong performance in agentic workflows over open-source models \citep{chang2024agentboard}, with GPT-4o-mini retaining competitive performance at a fraction of the inference cost \citep{xiao2024flowbench}. The framework is not restricted to any specific LLM, and alternative models (e.g. \textit{InternLM2.5-7B-Chat} \citep{cai2024internlm2}) may be substituted based on reasoning ability, code generation capability, and available compute.

\subsection{AnyChange for Zero-shot Change Detection}
\label{section:methodology-any-change}

AnyChange \citep{zheng2024segment} introduces a training-free framework for zero-shot change detection in RS imagery (Fig. \ref{fig:forest-chat-diagram} lower left, Fig. \ref{fig:any-change-diagram}), building on the Segment Anything Model (SAM) \citep{kirillov2023segment} without requiring fine-tuning or architectural modifications. It exploits the structured semantic properties of SAM's latent space to perform bi-temporal latent matching, comparing per-mask embeddings across temporal images using cosine similarity to identify semantically inconsistent regions. Temporal symmetry is ensured through bidirectional matching, with change predictions obtained by ranking confidence scores or thresholding on embedding angles. The model is designed to generalise across unseen change types and data distributions, producing either pixel-level or instance-level predictions. The model supports fully-automatic, semi-automatic (threshold-based), and point prompt-guided modes of operation.

To enable object-centric change detection, AnyChange combines bi-temporal latent matching with SAM's point prompt mechanism. This yields a \textit{point query mechanism} in which user-provided spatial coordinates are used to generate category-specific object proposals that are matched across time \citep{zheng2024segment}. Rather than requiring manual coordinate entry, we develop an interactive interface to allow users to click directly on areas of interest to generate point queries. The effect of point prompts on improving performance for a selected semantic class is well documented in \cite{zheng2024segment} and is not explored further here; for binary change detection, point prompts provide limited benefit given that change occurs within a single semantic class.

\subsection{\texorpdfstring{GPT-4o for Zero-shot Change Captioning and \\Refinement}{GPT-4o for Zero-shot Change Captioning and Refinement}}

Forest-Chat incorporates GPT-4o \citep{hurst2024gpt} for zero-shot change captioning and refinement, selected for its strong performance on single-image RS captioning \citep{ibrahimcomparative} and accessible multimodal API. Its application for RS change captioning remains largely unexplored \citep{soni2025earthdial}, though open-source alternatives also warrant further investigation \citep{leon2026describing}.

Bi-temporal image pairs are denormalised, base64-encoded, and passed to GPT-4o via the OpenAI multimodal API alongside a system prompt establishing an expert RS analyst role. Three prompting modes are evaluated. The \textit{general} prompt instructs the model to describe changes concisely in a single sentence. The \textit{style-guided} prompt provides few-shot caption examples and explicit style constraints to align outputs with the target caption distribution. The \textit{refinement} prompt operates as a two-stage pipeline in which FC-Supervised first generates a predicted caption, which is then passed alongside the bi-temporal image pair to GPT-4o, tasked with preserving the core change vocabulary while enriching the description with spatial grounding and domain-specific context directly observable in the imagery. A temperature of 0.2 is used across all modes to reduce output variability. A summary of the prompting strategies is provided in~\ref{appendix:prompts}. Full prompt templates are provided \href{https://github.com/JamesBrockUoB/ForestChat/blob/main/Multi_change/gpt4o_change_captioning.py}{here} for reproducibility.

\subsection{Multi-task Loss Balancing Strategy}
\label{section:multi-task-balancing-strategies}
During MCI model training, multi-task learning is performed jointly on change detection and change captioning, both using standard cross-entropy loss - applied to predicted change masks against ground truth annotations for detection, and to predicted sentences against ground truth descriptions for captioning. The two losses exhibit significant magnitude discrepancies, risking one task dominating the training signal. Additionally, conflicting gradients from the pixel-level detection and semantic captioning objectives can introduce destructive updates to shared representations during backpropagation.

To address this, we adopt a common multi-task loss balancing approach \citep{liu2024change} by normalising losses to the same order of magnitude with gradients detached for stability, ensuring equal task contribution and reducing the risk of one task dominating the optimisation process:

\begin{equation}
    \mathcal{L}_{\text{total}} = \frac{\mathcal{L}_{\text{det}}}{\text{detach}(\mathcal{L}_{\text{det}})} + \frac{\mathcal{L}_{\text{cap}}}{\text{detach}(\mathcal{L}_{\text{cap}})}
    \label{eq:original_loss_balancing}
\end{equation}

where detach($\cdot$) denotes the operation to detach the gradient, $\mathcal{L}_{\text{det}}$ is the change detection cross-entropy loss, and $\mathcal{L}_{\text{cap}}$ is the change captioning cross-entropy loss.

\subsection{Evaluation Metrics}
\label{section:experiments-metrics}

To evaluate the performance of Forest-Chat's supervised and zero-shot settings for binary change detection on RS imagery, the Mean Intersection over Union (mIoU) metric is employed, which quantifies segmentation quality by measuring the average overlap between predicted change masks and ground-truth masks across all classes:

\begin{equation}
\text{mIoU} = \frac{1}{C} \sum_{i=1}^{C} \frac{TP_i}{TP_i + FP_i + FN_i}
\end{equation}
where $C$ is the number of classes (i.e. change (\textit{c}), no change (\textit{nc})) for which metrics are calculated. For each class \textit{i}, $TP_i$ are true positives, $FP_i$ false positives, and $FN_i$ false negatives for class $i$.

By analysing both the individual class performances through IoU, as well as mIoU, model predictions for the under-represented change class can be compared. Overall accuracy (OA) is naturally a poor performance indicator for this task due to the severe class imbalances present in all three datasets. Despite LEVIR-MCI-Trees containing three classes (i.e. road, building, background), performance metrics are calculated on the basis of \textit{change (c)} and \textit{no change (nc)}, harmonising the results with those obtained for Forest-Change.

For evaluating the change captioning capabilities of FC-Supervised, FC-Zero-shot, and benchmarked models, common captioning evaluation metrics are applied for ease of comparison to the wider field. Additionally, BERTScore \citep{zhang2019bertscore} is employed to assess LLM-generated caption quality. The five metrics are summarised below:
\begin{itemize}
    \item \textbf{BLEU-n} \citep{papineni2002bleu}: Measures n-gram similarity between generated and reference sentences. Evaluated for n-grams of size 1 to 4, producing a total of four BLEU scores. Shortened to Bn in tables for readability.
    \item \textbf{METEOR} \citep{banerjee2005meteor}: Calculates the harmonic mean of unigram precision and recall, incorporating a penalty mechanism to assess generated sentence fluency.
    \item \textbf{ROUGE$_L$} \citep{lin2004rouge}: Measures similarity based on the longest common subsequence between generated and ground-truth sentences, prioritising recall performance for evaluating longer captions.
    \item \textbf{CIDEr-D} \citep{vedantam2015cider}: Computes cosine similarity between Term Frequency Inverse Document Frequency (TF-IDF) weighted n-gram vectors, treating each caption as a document to assess the semantic relevance.
    \item \textbf{BERTScore} \citep{zhang2019bertscore}: Computes semantic similarity between generated and reference captions using contextual embeddings from a pretrained transformer. By comparing cosine similarity between token embeddings rather than n-gram overlap, BERTScore captures paraphrasing and semantic equivalence, making it particularly suitable for evaluating LLM-generated captions.
\end{itemize}

Through these metrics, a comprehensive assessment of both change detection and captioning can be performed. By including per-class segmentation performance metrics, a deeper understanding of the under-represented \textit{change} class can be obtained. Across all metrics, higher scores indicate better change masks or improved sentence quality. 

\section{Experiments}
\label{section:experiments}

\subsection{Experimental Setup}
\label{section:exp-setup}
All models are implemented in PyTorch and trained via CPU on Isambard 3 \citep{green2025evaluation}, demonstrating that high computational resources are not required. Inference time benchmarking is conducted via CPU on an Apple M3 Pro. The maximum number of epochs is set to 100, with the backbone being trained until the sum of the mIoU and BLEU-4 scores does not improve for 10 consecutive epochs on the validation set. For FC-Supervised, once trained, the backbone network is subsequently frozen, and training is continued on the best model for the two branches separately. When training, the goal is to minimise the loss function defined in Equation \ref{eq:original_loss_balancing}, utilising the Adam optimiser with an initial learning rate of 0.0001.

\subsection{Forest-Chat Supervised Performance}
\label{section:experiments-mci-model-performance}

There remains a lack of available methods that simultaneously provide capabilities for joint change detection and captioning (CDC). Consequently, performance evaluations are done by comparing FC-Supervised to SOTA models for the two tasks separately on the LEVIR-MCI-Trees and Forest-Change datasets. Change3D \citep{zhu2025change3d} is capable of both tasks, but is not implemented in a unified manner. BiFA \citep{zhang2024bifa} is selected for change detection, and Chg2Cap \citep{chang2023changes} for change captioning. U-Net-based architectures remain popular for forest change segmentation \citep{nguyen2024multi}, and segmentation tasks more generally \citep{corley2024change}. We therefore benchmark against a U-Net SiamDiff model employing a pretrained ResNet-50 backbone \citep{he2016deep} with ImageNet \citep{deng2009imagenet} weights as a domain-specific baseline, following the implementation of \cite{corley2024change}. Additionally, differences in difficulty between the two datasets are identified by comparing performance within a low-data forest change analysis scenario against an urban-focused one derived from a comprehensive source dataset.

Table \ref{tab:combined-results} provides the performance comparisons for the benchmarked models on both  aforementioned datasets. FC-Supervised is the strongest performer for change captioning overall, and is the best on LEVIR-MCI-Trees for change detection. On the Forest-Change dataset, FC-Supervised is marginally outperformed by BiFA \cite{zhang2024bifa} for change class identification, with both models achieving similar overall performance. These results demonstrate strong generalisation capabilities across varying data domains for both change detection and captioning. The U-Net SiamDiff \cite{corley2024change} baseline was the worst performing change detection method on both datasets, particularly on the change class for Forest-Change. This highlights the value of designing architectures to explicitly process temporal information and task-aligned representations. The lightweight U-Net SiamDiff achieves the fastest change detection inference times, and the use of pretrained weights substantially improves performance compared to the non-pretrained Siamese Attention U-Net reported by \cite{hewarathna2024change} on the Forest-Change dataset (73.78\% F1 vs 42.84\% F1), from which our data is derived. Among the benchmarked methods, FC-Supervised is the slowest at captioning inference and the second slowest for change detection (see Table \ref{tab:inference-time}). This is expected because it performs joint multi-task reasoning with heavy cross-modal and cross-temporal interactions and multi-task attention. However, these inference times are not prohibitive for practical applications, especially given that only a single model is required to generate pixel-level and semantic-level information of changes within RS image pairs. Zero-shot inference times are discussed in Section \ref{section:experiments-any-change-performance}.

\begin{table*}[ht!]
\centering
\caption{Change Detection and Change Captioning performances on the test sets of LEVIR-MCI-Trees and Forest-Change datasets Best results per dataset and metric are \textbf{bold}, second-best are \underline{underlined}. Results are an average of three runs.}
\label{tab:combined-results}
\setlength{\tabcolsep}{4pt}

\begin{tabular}{c | c | c c c | c c c c c c c}
\toprule
Dataset & Model & mIoU & $IoU_{\textit{nc}}$ & $IoU_{\textit{c}}$ & B1 & B2 & B3 & B4 & METEOR & ROUGE$_L$ & CIDEr-D \\
\midrule
\multirow{5}{*}{LEVIR-MCI-Trees}
 & U-Net SiamDiff & 86.08 & 95.04 & 77.11 & - & - & - & - & - & - & - \\
 & BiFA
 & \underline{87.54} & \underline{95.63} & \underline{79.45} 
 & - & - & - & - & - & - & - \\

 & Chg2Cap
 & - & - & - 
 & \underline{70.25} & 53.88 & \underline{38.56} & \underline{27.28} & \underline{21.80} & 45.65 & \underline{37.72} \\

 & Change3D
 & 87.48 & \underline{95.63} & 79.34 
 & 69.52 & \underline{54.35} & 38.33 & 26.41 & 21.57 & \underline{46.83} & 35.03 \\

 & FC-Supervised
 & \textbf{88.13} & \textbf{95.89} & \textbf{80.36} 
 & \textbf{75.25} & \textbf{60.90} & \textbf{46.21} & \textbf{34.41} & \textbf{23.32} & \textbf{49.34} & \textbf{48.69} \\

\cmidrule(l){1-12}
\multirow{5}{*}{Forest-Change}
 & U-Net SiamDiff & 64.42 & 95.71 & 33.14 & - & - & - & - & - & - & - \\
 & BiFA
 & \textbf{67.34} & 95.85 & \textbf{38.84} 
 & - & - & - & - & - & - & - \\

 & Chg2Cap
 & - & - & - 
 & 59.09 & 45.12 & 34.59 & 27.10 & 23.50 & 43.41 & 12.97 \\

 & Change3D
 & 66.01 & \underline{95.93} & 36.08 
 & \underline{61.08} & \underline{49.18} & 40.46 & 33.32 & 25.65 & \underline{46.26} & 20.78 \\

 & FC-Supervised
 & \underline{67.10} & \textbf{96.12} & \underline{38.07} 
 & \textbf{67.54} & \textbf{56.34} & \textbf{47.55} & \textbf{40.17} & \textbf{28.22} & \textbf{48.52} & \textbf{38.79} \\

\bottomrule
\end{tabular}
\end{table*}

\begin{table}[ht!]
\centering
\caption{Inference time per iteration (s), averaged across LEVIR-MCI-Trees and Forest-Change datasets.}
\label{tab:inference-time}
\begin{tabular}{c | c | c}
\toprule
\textbf{Task} & \textbf{Model} & \textbf{Time / iter (s)} \\
\midrule
\multirow{4}{*}{Change Captioning}
    & FC-Zero-shot & \textbf{3.34*}\\
    & Chg2Cap & 3.81\\
    & Change3D & 4.00 \\
    & FC-Supervised & 5.35 \\
\midrule
\multirow{7}{*}{Change Detection}
    & U-Net SiamDiff & \textbf{0.12} \\
    & Change3D & 0.53 \\
    & BiFA & 0.84 \\
    & FC-Supervised & 1.84 \\
    & FC-Zero-shot$_{ViT-B}$ & 16.07 \\
    & FC-Zero-shot$_{SAM2}$ & 23.28 \\
    & FC-Zero-shot$_{ViT-H}$ & 31.88 \\
    
\bottomrule
\end{tabular}
\footnotesize{*GPT-4o inference within FC-Zero-shot includes API communication overhead and is not directly comparable to local inference times.}
\end{table}

Overall change detection performance is considerably better for building and road detection than for deforestation, likely due to the pronounced class imbalances within Forest-Change, fuzzy edges at the borders of many deforestation patches, the uniqueness of deforestation patterns, and the relatively small dataset size. Within the Forest-Change dataset, a considerable percentage of change masks are comprised of numerous tiny patches that are challenging for models to detect. This resulted in all benchmarked CD models frequently predicting change masks that mirrored the deforestation patterns observed, but contained few overlaps with the ground-truth masks (see row 4 in Fig. \ref{fig:qualitative_performance}). Models generally perform well on larger patches, but can still struggle with accurately predicting fuzzy boundaries. This corroborates the findings of \cite{liu2024change}, who highlight that the segmentation performance of objects containing fewer than 400 pixels is low. Compared with deforestation detection, building change detection patterns and object geometries are more predictable, further raising the challenge posed by the Forest-Change dataset. 

Captioning performance evaluation is more nuanced. For the LEVIR-MCI-Trees dataset, the BLEU-1, BLEU-2, and CIDEr-D metrics are generally higher, while BLEU-3, BLEU-4, METEOR, and ROUGE$_L$ are better for Forest-Change. The observed results largely reflect the differences in annotation styles between the two datasets, as described in Section \ref{section:data-caption-generation}. Despite models frequently predicting deforestation severity accurately, they occasionally get confused about the location and characterisation of deforestation patches, and are unable to recognise geographic features within the imagery. To tackle this shortcoming, we investigate caption refinement through domain-aware zero-shot MLLMs in Section \ref{section:experiments-llm-caption-generation-refinement}. Alternative domain-aware strategies, such as vocabulary expansion \citep{gao2024ve} or domain-adaptive fine-tuning \citep{guo2022clip4idc}, could also help address these limitations.

\subsection{Forest-Chat Zero-shot Change Detection}
\label{section:experiments-any-change-performance}

In keeping with the previous section, FC-Zero-shot performance is evaluated on both the Forest-Change and LEVIR-MCI-Trees datasets. To explore the impact of different backbone architectures, three SAM-based configurations are evaluated: the default \textit{ViT-H} backbone used in the Segment Any Change framework \citep{zheng2024segment}, a smaller \textit{ViT-B} backbone, and the more recent SAM2 \citep{ravi2024sam} architecture. These alternatives provide insight into the trade-offs between segmentation performance and inference efficiency for zero-shot change detection.

To optimise performance, a Bayesian hyperparameter search comprising 20 runs for each backbone was conducted to identify the most influential parameters. Following the Segment Any Change framework \citep{zheng2024segment}, bi-temporal matching and feature normalisation are enabled throughout due to their consistently positive impact on change discrimination. The remaining hyperparameters are not tuned due to their minimal impact on mask generation. Points-per-side - which controls the spatial density of point prompts provided to SAM for mask proposal generation - is fixed at 16 as a practical trade-off between performance and inference time. Performing a full hyperparameter sweep on LEVIR-MCI-Trees is computationally prohibitive; therefore, the best configuration from Forest-Change is reused. These parameters mainly control proposal density and filtering heuristics, rather than dataset-specific semantics, so reusing them is unlikely to introduce systematic bias.

The following hyperparameters are tuned:

\begin{itemize}
\item \textbf{Change confidence threshold}: Specifies the minimum cosine distance in the latent feature space required to classify a region as changed, directly governing the trade-off between sensitivity and false positives.
\item \textbf{Stability score threshold}: Filters mask proposals based on their consistency under point prompt perturbations, suppressing unstable or noisy change predictions.
\item \textbf{Area threshold}: Suppresses small or fragmented change regions by enforcing a minimum relative area requirement, helping reduce spurious detections in cluttered scenes.
\item \textbf{Object similarity threshold}: Regulates the tolerance for matching corresponding regions across temporal images, influencing the balance between over-segmentation and missed change instances.
\end{itemize}

The resulting segmentation performance for each backbone and configuration is summarised in Table \ref{tab:anychange-comparison}. Among the evaluated hyperparameters, the change confidence threshold has the greatest influence on performance, as it directly controls the sensitivity of the cosine-distance based change detection mechanism. Across the evaluated backbones, hyperparameter tuning consistently improves segmentation quality across both classes by increasing sensitivity to genuine change regions while suppressing spurious detections caused by atmospheric artifacts and seasonal variations.

Zero-shot performance remains substantially lower than the supervised FC-Supervised baseline across both datasets, with the gap particularly pronounced on the LEVIR-MCI-Trees dataset. This discrepancy can partly be attributed to the dataset’s annotation protocol, which only labels road and building changes. Consequently, other legitimate changes detected by the model are penalised during evaluation, contributing to the dataset’s observed sensitivity to hyperparameter selection. In contrast, the Forest-Change dataset adopts a more consistent definition of change, resulting in comparatively stronger zero-shot performance and greater robustness to hyperparameter variation.

\begin{table}[ht!]
\centering
\caption[Zero-shot Segmentation Results]{FC-Zero-shot segmentation performance with default parameters and best hyperparameter sweep configuration for the SAM (\textit{ViT-B} and \textit{ViT-H}) and SAM2 (\textit{SAM 2.1 Hiera Large}) models. SAM Best configuration : change confidence threshold = 145; stability score threshold = 0.93; area threshold = 0.9; object similarity threshold = 60. SAM2 Best configuration : change confidence threshold = 150; stability score threshold = 0.91; area threshold = 0.9; object similarity threshold = 50. P = precision, R = recall.}
\label{tab:anychange-comparison}
\setlength{\tabcolsep}{4pt}
\scriptsize

\begin{tabular}{c | c | c c c c c }
\toprule
Dataset & Model & mIoU & $IoU_{\textit{nc}}$ & $IoU_{\textit{c}}$ & P & R \\
\midrule
\multirow{7}{*}{LEVIR-MCI-Trees}
 & FC-Zero-shot$_{ViT-B}^{D}$ & 44.55 & 72.80 & 16.29 & 55.84 & 56.68 \\
 & FC-Zero-shot$_{ViT-B}^{B}$ & 48.88 & 76.20 & 21.56 & 60.58 & 61.49 \\
 & FC-Zero-shot$_{ViT-H}^{D}$ & 38.04 & 54.05 & 22.04 & 57.51 & 63.65 \\
 & FC-Zero-shot$_{ViT-H}^{B}$ & 47.32 & 66.60 & 28.04 & 62.08 & 70.34 \\
 & FC-Zero-shot$_{SAM2}^D$ & 39.07 & 57.32 & 20.82 & 56.51 & 61.58 \\
 & FC-Zero-shot$_{SAM2}^B$ & 48.12 & 74.55 & 21.69 & 59.85 & 61.72 \\
 \cmidrule(l){2-7}
 & FC-Supervised & 88.13 & 95.89 & 80.36 & 94.30 & 92.92 \\
\cmidrule(l){1-7}
\multirow{7}{*}{Forest-Change}
 & FC-Zero-shot$_{ViT-B}^{D}$ & 58.11 & 95.13 & 21.09 & 75.80 & 62.15 \\
 & FC-Zero-shot$_{ViT-B}^{B}$ & 59.24 & 95.51 & 22.97 & 80.93 & 62.69 \\
 & FC-Zero-shot$_{ViT-H}^{D}$ & 58.56 & 93.83 & 23.29 & 67.97 & 66.69 \\
 & FC-Zero-shot$_{ViT-H}^{B}$ & 60.15 & 95.11 & 25.19 & 75.26 & 65.30 \\
 & FC-Zero-shot$_{SAM2}^D$ & 56.62 & 94.92 & 18.40 & 73.41 & 60.62 \\
 & FC-Zero-shot$_{SAM2}^B$ & 58.16 & 93.92 & 22.40 & 68.01 & 65.66 \\
 \cmidrule(l){2-7}
 & FC-Supervised & 67.10 & 96.12 & 38.07 & 84.19 & 76.82 \\
\bottomrule
\end{tabular}
\begin{center}
\footnotesize{$^{D}$ Default parameters;\quad $^{B}$ Best hyperparameter configuration.}
\end{center}

\end{table}

Backbone selection introduces a clear trade-off between segmentation quality, detection sensitivity, and computational cost. As shown in Table \ref{tab:anychange-comparison}, the larger \textit{ViT-H} backbone generally achieves stronger $IoU_{\textit{c}}$ performance, indicating improved sensitivity to subtle or fragmented change regions. However, this increased sensitivity can also amplify false detections caused by illumination shifts, cloud cover, or phenological differences between image acquisition dates, as illustrated qualitatively in Figure \ref{fig:qualitative_performance}. This behaviour is reflected in the precision–recall characteristics reported in Table \ref{tab:anychange-comparison}. On the LEVIR-MCI-Trees dataset, the best-performing \textit{ViT-H} configuration achieves higher recall (70.34\%) than \textit{ViT-B} (61.49\%), indicating greater sensitivity to change regions, but with only marginal improvements in precision. A similar trend is observed on the Forest-Change dataset, where \textit{ViT-H} attains higher recall (66.69\%) but lower precision (67.97\%) relative to the more conservative \textit{ViT-B} configuration, which achieves higher precision (80.93\%) at the cost of reduced recall (62.69\%).

The SAM2 backbone provides an intermediate alternative, achieving comparable overall performance while partially reducing the computational burden relative to larger encoders such as \textit{ViT-H}. However, $IoU_{\textit{c}}$ performance remains comparatively limited. Although SAM2 introduces architectural improvements designed to improve efficiency and temporal consistency in video segmentation, these advantages are not fully utilised in the present pipeline, which processes the two temporal images independently and relies primarily on feature similarity for change inference. Consequently, the SAM2 configuration does not consistently outperform the original Segment Anything Model backbones in this setting.

Despite lowering the points-per-side hyperparameter, inference time remains substantially slower than other benchmarked methods (Table \ref{tab:inference-time}), limiting practical deployment in large-scale monitoring scenarios. While smaller backbones can partially reduce this computational burden, the associated decline in segmentation quality limits the practical benefits of this trade-off. In practice, improvements in segmentation sensitivity do not always correspond to cleaner change maps, particularly when radiometric, atmospheric, or seasonal variations are interpreted as genuine change.

These results suggest that purely zero-shot approaches may have inherent limitations for reliable change detection in remote sensing imagery. Rather than relying exclusively on zero-shot inference, future work may benefit from exploring architectures that combine strong pretrained encoders with lightweight adaptation mechanisms capable of learning from limited labelled data. Approaches such as META-CD \cite{gao2025combining} demonstrate that applying domain-adaptive pretraining for vision foundation models can significantly improve change detection performance while maintaining strong generalisation under limited supervision, and lighter backbones could be selected to balance performance against inference times.

\subsection{
\texorpdfstring{Forest-Chat Zero-shot for Change Captioning\\and Refinement}{Forest-Chat Zero-shot for Change Captioning and Refinement}}
\label{section:experiments-llm-caption-generation-refinement}

\begin{table*}[ht!]
\centering
\caption{FC-Zero-shot Change Captioning and Caption Refinement on the test sets of LEVIR-MCI-Trees and Forest-Change datasets. Results for refined captions are an average of three runs.}
\label{tab:llm-caption-generation-refinement}
\setlength{\tabcolsep}{4pt}

\begin{tabular}{c | c | c c c c c c c c}
\toprule
Dataset & Model & B1 & B2 & B3 & B4 & METEOR & ROUGE$_L$ & CIDEr-D & BERTScore \\
\midrule
\multirow{4}{*}{LEVIR-MCI-Trees}
 & FC-Zero-shot$^G$ & 52.29 & 31.90 & 17.89 & 10.19 & 18.00 & 32.94 & 18.86 & 86.42 \\

 & FC-Zero-shot$^S$ & 67.56 & 45.68 & 29.45 & 18.23 & 21.19 & 37.20 & 29.95 & 84.27 \\

 & FC-Supervised
 & 75.25 & 60.90 & 46.21 & 34.41 & 23.32 & 49.34 & 48.69 & 86.34 \\

 & FC-Supervised-Refined & 59.49 & 40.57 & 27.14 & 17.54 & 22.90 & 38.81 & 18.49 & 87.36 \\

\cmidrule(l){1-10}
\multirow{4}{*}{Forest-Change} 
 & FC-Zero-shot$^G$
 & 25.88 & 10.86 & 3.65 & 0.00 & 8.99 & 21.11 & 2.06 & 82.82 \\

 & FC-Zero-shot$^S$ & 52.97 & 43.64 & 38.20 & 34.00 & 24.96 & 39.87 & 32.13 & 82.47 \\

  & FC-Supervised
 & 67.54 & 56.34 & 47.55 & 40.17 & 28.22 & 48.52 & 38.79 & 85.63 \\

 & FC-Supervised-Refined & 73.19 & 59.87 & 50.63 & 43.31 & 32.98 & 54.19 & 50.83 & 86.73 \\

\bottomrule
\end{tabular}
\begin{center}
\footnotesize
$^{G}$ General prompt for FC-Zero-shot;\quad
$^{S}$ Dataset-specific style-guided prompt for FC-Zero-shot.
\end{center}
\end{table*}

Two prompting strategies are compared: a general prompt providing minimal task context, and a dataset-specific style-guided prompt designed to align FC-Zero-shot's outputs with the vocabulary, structure, and thematic content of each dataset. As shown in Table~\ref{tab:llm-caption-generation-refinement}, style-guided prompting substantially improves zero-shot performance across both datasets, with BLEU-4 increasing from 10.19 to 18.23 on LEVIR-MCI-Trees and from 0.00 to 34.00 on Forest-Change. The latter result highlights that zero-shot general prompting alone is insufficient to produce captions consistent with the semantic and stylistic conventions of the target dataset. Despite these gains, FC-Zero-shot under style-guided prompting remains below FC-Supervised on most metrics, confirming that zero-shot MLLMs cannot yet supplant supervised change captioning models in specialised remote sensing domains.

While zero-shot performance is limited, FC-Zero-shot can still provide value as a targeted refinement stage for FC-Supervised outputs, with the extent of benefit varying between datasets. On Forest-Change, refinement consistently improves captions. The reference captions are largely rule-based, which causes much of the semantic detail available in the human-written captions to be overlooked. Refinement enriches the captions with additional spatial and contextual information, which is reflected in improvements in BLEU-4 (40.17 → 43.31) and CIDEr-D (38.79 → 50.83), while preserving the underlying change semantics. In contrast, refinement on LEVIR-MCI-Trees introduces additional descriptive detail that, while semantically accurate, diverges lexically from the concise human-written references. This results in declines in n-gram metrics (BLEU-4: 34.41 -> 17.54, CIDEr-D: 48.69 -> 18.49) while BERTScore improves slightly (86.34 → 87.36). This demonstrates that semantic content is preserved even when lexical overlap decreases. These results indicate that refinement is practically valuable as a targeted post-processing step for known supervised model limitations (e.g. correcting geographic positioning, fixing grammatical inconsistencies, or incorporating domain-specific  vocabulary), rather than as a general-purpose quality enhancement. Tables~\ref{tab:refinement-forest} and~\ref{tab:refinement-levir} in the appendix present selected examples illustrating the spatial and semantic enrichment introduced by FC-Zero-shot refinement over FC-Supervised predicted captions on the Forest-Change and LEVIR-MCI-Trees test sets respectively.

Overall, these results show that multimodal LLMs can be effective  for change captioning, particularly when guided by dataset-specific prompts or used for targeted refinement. Zero-shot utility remains limited without explicit domain grounding, and the benefits of refinement are dependent on the reference caption characteristics. Forest-Chat's innate supervised captioning remains strong, but the addition of zero-shot captioning and refinement offers significant utility to users. Future work should broaden benchmarking of open-source models within the domain, and investigate fine-tuning strategies or incorporating domain-aware knowledge bases to better align LLM outputs with the structural and vocabulary constraints of remote sensing change captioning datasets.

\subsection{Cross-Domain Adaptation Study}
\label{section:domain-transfer}
Assessing model generalisation across diverse remote sensing scenes and sensors is critical for understanding downstream performance and real-world operational behaviour. Consequently, this section evaluates the change detection models benchmarked in Section \ref{section:experiments-mci-model-performance} by first pretraining on Forest-Change and then fine-tuning on varying proportions of the JL1-CD-Trees training set, from 5\% up to 100\%, before evaluation on its test set. FC-Zero-shot serves as a training-free baseline. For FC-Supervised, we additionally compare pretraining on Forest-Change using either full multi-task learning or change-detection-only to assess the transferability of task-specific versus multi-task features. We also evaluate each model when trained on the full JL1-CD-Trees dataset with and without Forest-Change pretraining, to directly quantify how effectively pretrained features improve performance in the new domain.

\begin{figure*}[ht!]
    \centering
    \includegraphics[width=1\linewidth]{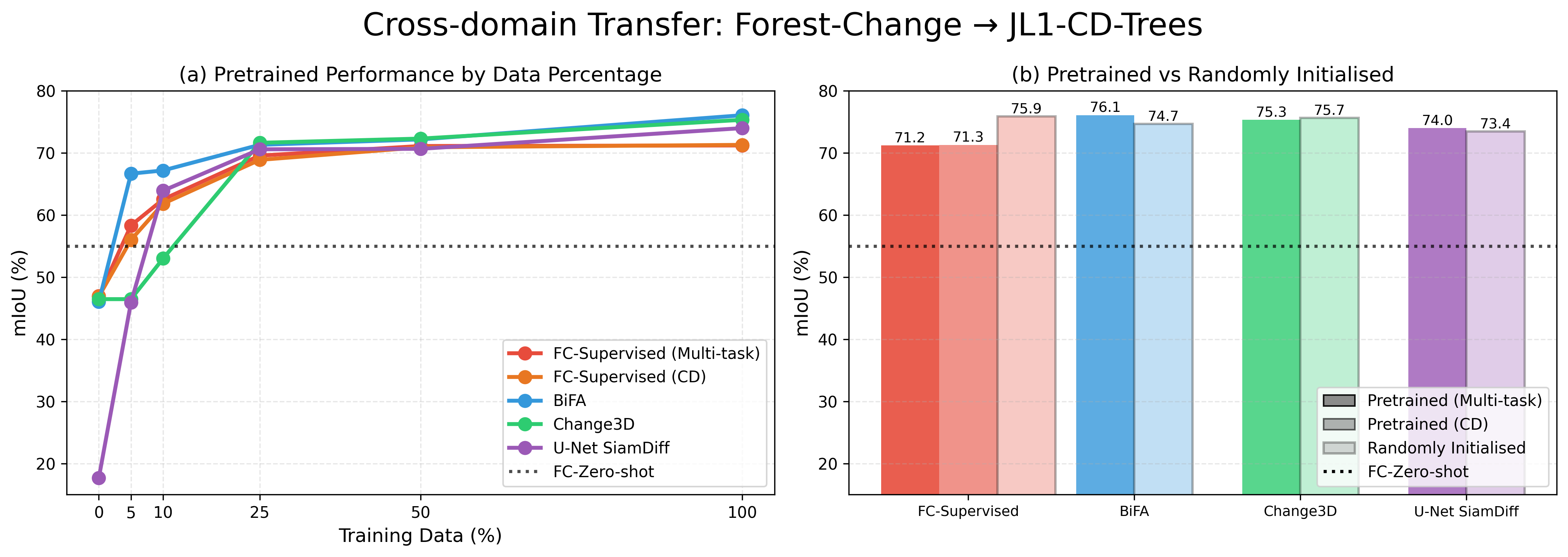}
    \caption[Cross-Domain Transfer]{Cross-domain transfer performance on JL1-CD-Trees dataset. a) Models pretrained on Forest-Change and fine-tuned on increasing fractions of JL1-CD-Trees. b) Comparison of models trained solely on JL1-CD-Trees versus pretrained on Forest-Change and fine-tuned on the full train set. FC-Zero-shot is included as a baseline.}
    \label{fig:cross-domain-transfer}
\end{figure*}

The performance trends observed on the JL1-CD-Trees dataset in Figure \ref{fig:cross-domain-transfer} reveal clear patterns in how the benchmarked architectures adapt to cross-resolution target-domain data. Zero-shot transfer from Forest-Change pretraining yields consistently poor performance across all supervised models, with a particularly severe collapse for U-Net SiamDiff. In contrast, FC-Zero-shot achieves an mIoU of 54.98 using Forest-Change-optimised hyperparameters, demonstrating consistent cross-domain change detection capabilities under training-free conditions. The near-uniform collapse in zero-shot performance across supervised models underscores the severity of the domain gap between Forest-Change and JL1-CD-Trees, driven most consequentially by the order-of-magnitude difference in spatial resolution (30 m vs <1 m) and the absence of matched seasonal representation in the source domain.

However, once target-domain supervision is introduced, performance improves rapidly across all supervised architectures. Even with only 5–10\% of the JL1-CD-Trees training data, mIoU rises substantially from zero-shot levels, where most models initially default to predicting no change, or behave near-randomly as in the case of U-Net SiamDiff. This indicates that while pretrained features provide a useful starting point, models require even modest target-domain supervision to correctly interpret high-resolution change patterns. Performance continues to increase steadily as more data is added, with most architectures approaching mIoU values in the low-to-mid 70s when trained on the full dataset. Every model exhibits a notable performance jump between 10\% and 25\% of the training data, after which improvements become more incremental, suggesting that a relatively small amount of target-domain supervision is sufficient to recalibrate models to the spatial characteristics of the JL1-CD-Trees imagery, with additional data primarily refining boundary precision and small-object detection.

The role of Forest-Change pretraining is more nuanced. While pretraining accelerates early-stage adaptation, it does not consistently improve final performance when the full JL1-CD-Trees training set is available. In several cases, models trained from scratch on JL1-CD-Trees achieve comparable or slightly higher performance than their pretrained counterparts. FC-Supervised provides a notable example, achieving stronger results when trained directly on JL1-CD-Trees than when initialised with Forest-Change pretraining, suggesting that coarse-resolution representations may introduce biases that hinder adaptation to the finer spatial structures present in the target dataset. In contrast, architectures such as BiFA and Change3D benefit more consistently from increasing supervision when initialised with Forest-Change pretraining, with BiFA achieving the strongest performance among the pretrained models when trained on the full dataset. Differences between the multi-task and CD-only FC-Supervised variants are comparatively small, indicating that the additional captioning supervision does not substantially alter transfer behaviour for the detection task. Overall, these findings suggest that while pretraining on related RS datasets can provide useful initialisation, its effectiveness depends strongly on the quality of the source dataset, and the spatial alignment between source and target domains.

\subsection{Ablation Studies}
\subsubsection{\texorpdfstring{Multi-task Loss Balancing and Gradient Conflict \\Resolution Study}{Multi-task Loss Balancing and Gradient Conflict Resolution Study}}
\label{section:experiments-loss-balancing}

As discussed in Section~\ref{section:multi-task-balancing-strategies}, joint optimisation of change detection and captioning may introduce gradient interference due to the differing objectives of spatial localisation and semantic language generation. While normalised loss balancing is effective at preventing one task from dominating the learning gradient, it assumes that tasks are equally important and difficult throughout training, potentially ignoring the fundamental noise heteroscedasticity between tasks. To investigate this, we experiment with various multi-task learning (MTL) loss balancing and gradient conflict resolution techniques to assess their impact on the joint training of CDC and their potential to improve collaborative learning. Several commonly used gradient adjustment methods are compared to examine their influence on multi-task optimisation.

For MTL loss balancing, we select Uncertainty Weighting \citep{kendall2018multi} and Enhanced Dynamic Weight Averaging (EDWA) \citep{shi2024multi}. For gradient conflict resolution, we evaluate Conflict-Averse Gradient Descent (CAGrad) \citep{liu2021conflict}, Projecting Conflicting Gradients (PCGrad) \citep{yu2020gradient}, and GradDrop \citep{chen2020just}. To ensure a comprehensive experiment suite, every methodological combination is explored.

\begin{table*}[ht!]
\centering
\ssmall
\caption[Aggregated Change Detection and Change Captioning Performance]{\textbf{Aggregated change detection and change captioning performance for MTL / Gradient Conflict resolution strategies.} Mean $\pm$ standard deviation across configurations, with three runs each. Bold indicates the best mean value per dataset per metric within each method group. When results are tied or close, preference is given to the method with the lower standard deviation.}
\label{tab:mtl-strategy-results}
\setlength{\tabcolsep}{4pt} 

\begin{tabular}{c | c | c c c | c c c c c c c}
\toprule
\textbf{Dataset} & \textbf{MTL / Gradient Conflict Strategy} & mIoU & $IoU_{\textit{nc}}$ & $IoU_{\textit{c}}$ & B1 & B2 & B3 & B4 & METEOR & ROUGE$_L$ & CIDEr-D \\
\midrule
\multirow{7}{*}{LEVIR-MCI-Trees}
 & Equal & \textbf{88.08$\pm$0.15} & \textbf{95.88$\pm$0.04} & \textbf{80.28$\pm$0.27} & 74.66$\pm$1.34 & 59.88$\pm$1.27 & 44.92$\pm$1.28 & 33.12$\pm$1.19 & \textbf{23.48$\pm$0.65} & 49.75$\pm$0.89 & 48.05$\pm$2.80 \\
 & EDWA        & 87.17$\pm$0.55 & 95.53$\pm$0.18 & 78.80$\pm$0.94 & 72.22$\pm$4.06 & 57.68$\pm$3.74 & 42.31$\pm$4.38 & 30.60$\pm$4.07 & 22.44$\pm$1.65 & 48.34$\pm$1.79 & 43.19$\pm$7.14 \\
 & Uncertainty & 87.51$\pm$0.48 & 95.64$\pm$0.18 & 79.38$\pm$0.78 & \textbf{74.72$\pm$1.37} & \textbf{60.13$\pm$1.38} & \textbf{45.46$\pm$1.55} & \textbf{33.82$\pm$1.85} & 23.41$\pm$0.55 & \textbf{49.89$\pm$0.81} & \textbf{48.80$\pm$3.13} \\
\cmidrule(lr){2-12}
 & None     & \textbf{87.85$\pm$0.32} & \textbf{95.77$\pm$0.12} & \textbf{79.93$\pm$0.52} & \textbf{74.80$\pm$1.46} & \textbf{60.24$\pm$1.12} & \textbf{45.36$\pm$1.40} & \textbf{33.52$\pm$1.71} & \textbf{23.33$\pm$0.62} & 49.40$\pm$0.67 & \textbf{47.44$\pm$3.29} \\
 & CAGrad   & 87.80$\pm$0.62 & 95.77$\pm$0.21 & 79.83$\pm$1.04 & 73.51$\pm$3.85 & 58.76$\pm$3.83 & 43.93$\pm$4.45 & 32.44$\pm$4.27 & 23.01$\pm$1.61 & 49.18$\pm$2.18 & 47.43$\pm$6.68 \\
 & PCGrad   & 87.38$\pm$0.62 & 95.60$\pm$0.24 & 79.16$\pm$0.99 & 73.24$\pm$3.31 & 58.30$\pm$2.76 & 43.19$\pm$3.46 & 31.57$\pm$3.11 & 22.91$\pm$1.51 & 49.10$\pm$1.61 & 45.48$\pm$6.58 \\
 & GradDrop & 87.31$\pm$0.53 & 95.60$\pm$0.20 & 79.02$\pm$0.87 & 73.92$\pm$2.08 & 59.63$\pm$1.88 & 44.44$\pm$2.00 & 32.52$\pm$2.25 & 23.19$\pm$0.60 & \textbf{49.62$\pm$0.80} & 46.38$\pm$4.53 \\
\midrule
\multirow{7}{*}{Forest-Change}
 & Equal      & \textbf{67.34$\pm$0.44} & \textbf{96.20$\pm$0.12} & \textbf{38.47$\pm$0.85} & \textbf{64.68$\pm$3.95} & 53.88$\pm$4.12 & 45.33$\pm$4.41 & 38.43$\pm$4.52 & 27.55$\pm$1.87 & 47.55$\pm$2.74 & \textbf{35.57$\pm$8.53} \\
 & EDWA       & 66.88$\pm$0.70 & 96.19$\pm$0.20 & 37.56$\pm$1.35 & 64.46$\pm$4.47 & 53.69$\pm$4.29 & 44.78$\pm$4.36 & 37.73$\pm$4.53 & 27.47$\pm$2.60 & 47.71$\pm$3.21 & 33.13$\pm$6.35 \\
 & Uncertainty & 67.02$\pm$0.55 & 96.22$\pm$0.31 & 37.83$\pm$0.95 & 63.89$\pm$3.41 & \textbf{53.88$\pm$3.49} & \textbf{45.65$\pm$3.69} & \textbf{39.10$\pm$4.03} & \textbf{27.95$\pm$2.15} & \textbf{48.67$\pm$2.90} & 33.74$\pm$8.09 \\
\cmidrule(lr){2-12}
 & None     & \textbf{67.23$\pm$0.34} & 96.24$\pm$0.23 & \textbf{38.22$\pm$0.53} & \textbf{66.00$\pm$4.37} & \textbf{55.51$\pm$3.75} & \textbf{46.99$\pm$3.57} & \textbf{40.05$\pm$3.67} & \textbf{28.34$\pm$2.41} & \textbf{48.73$\pm$3.16} & \textbf{36.84$\pm$9.95} \\
 & CAGrad   & 67.16$\pm$0.53 & \textbf{96.26$\pm$0.16} & 38.06$\pm$0.92 & 63.68$\pm$2.49 & 53.49$\pm$3.06 & 45.47$\pm$3.69 & 39.00$\pm$4.22 & 27.77$\pm$1.26 & 48.40$\pm$1.98 & 35.05$\pm$6.47 \\
 & PCGrad   & 66.87$\pm$0.70 & 96.19$\pm$0.13 & 37.54$\pm$1.42 & 64.39$\pm$3.71 & 53.72$\pm$3.57 & 44.95$\pm$3.58 & 38.13$\pm$3.84 & 27.25$\pm$2.10 & 47.15$\pm$2.60 & 35.03$\pm$5.31 \\
 & GradDrop & 67.05$\pm$0.75 & 96.11$\pm$0.31 & 37.99$\pm$1.39 & 63.29$\pm$4.64 & 52.54$\pm$4.91 & 43.60$\pm$5.14 & 36.48$\pm$5.13 & 27.25$\pm$2.83 & 47.62$\pm$3.84 & 29.65$\pm$6.97 \\
\bottomrule
\end{tabular}
\end{table*}

Experiments are performed across the Forest-Change and LEVIR-MCI-Trees datasets to assess sensitivity to data domain. Due to the stochastic nature of the full experimental results presented in Tables ~\ref{tab:mtl-ablation-results-levir}--\ref{tab:mtl-ablation-delta-fc}, the results for each method are aggregated and presented in Table \ref{tab:mtl-strategy-results}. Across both datasets, segmentation performance remains largely stable regardless of the multi-task learning optimisation strategy, with Equal weighting only having an mIoU improvement of \textasciitilde0.3-0.5 over Uncertainty. In contrast, captioning metrics exhibit greater variability, indicating that multi-task interactions primarily affect the language generation component. Here, Uncertainty is generally the better-performing MTL loss balancing method, with BLEU-4 \textasciitilde0.7 greater than Equal losses. For the majority of metrics, no gradient conflict resolution techniques are required to stabilise task harmony. The only exception to this is Table \ref{tab:mtl-ablation-delta-levir}, in which Uncertainty with CAGrad on the LEVIR-MCI-Trees dataset displays notable improvements for both tasks, but this result is not replicated on Forest-Change. The observed invariance of change detection performance and the sensitivity of change captioning to MTL loss balancing strategy is likely due to the dense supervision and complementary nature of the segmentation task with FC-Supervised, which stabilises shared feature representations. Conversely, the captioning task receives sparser, sequential supervision and is more sensitive to variations in gradient contributions. These results suggest that, in settings where tasks are naturally aligned, careful architectural design and shared representations may be more important than sophisticated multi-task optimisation strategies. In such cases, loss balancing and gradient conflict resolution can introduce unnecessary optimisation noise, potentially disrupting beneficial cross-task regularisation. Developing architectures that allow tasks to reinforce each other during joint learning should therefore be prioritised over relying on complex optimisation techniques. Future work could also explore meta-learning based task weighting approaches, such as Auto-Lambda \citep{liu2022auto}, which dynamically learn task weights through meta-optimisation and may further improve multi-task synergy.

\subsubsection{Forest Chat Supervised Backbone Size Influence}
\label{section:experiments-backbone-size}

This section reports on the importance of the Segformer backbone model for feature extraction on downstream tasks. By default, a MiT-B1 is used, but this study evaluates change detection and captioning performance using the smaller MiT-B0 and larger MiT-B2 variants. To provide a lightweight convolutional comparison, a RegNetX-400MF \citep{radosavovic2020designing} backbone is also evaluated. Due to computational limitations and the diminishing returns observed in the original SegFormer paper \citep{xie2021segformer}, backbone sizes above B2 are not explored in this study. Table \ref{tab:backbone-results} displays a general trend of increasing change detection performance for both the LEVIR-MCI-Trees and Forest-Change datasets. FC-Supervised with the MiT-B2 backbone achieves the strongest change detection performance across both datasets, with larger backbone sizes improving change class identification. On the Forest-Change dataset, the MiT-B2 configuration also outperforms BiFA \citep{zhang2024bifa} in Table \ref{tab:combined-results}, making FC-Supervised the best performing model when using a larger backbone.

\begin{table*}[ht!]
\centering
\small
\caption{Change Detection and Change Captioning performances on the test sets of LEVIR-MCI-Trees and Forest-Change datasets using different Segformer (MiT-Bn) backbone sizes and RegNetX-400MF for FC-Supervised. Best results per metric are \textbf{bold}, second-best are \underline{underlined}. The training time is reported as a rounded mean $\pm$ standard deviation (s) for a single epoch. Results are averages of three runs.}
\label{tab:backbone-results}
\setlength{\tabcolsep}{3pt}

\begin{tabular}{c | c | c c c | c c c c c c c | c}
\toprule
Dataset & Backbone & mIoU & $IoU_{\textit{nc}}$ & $IoU_{\textit{c}}$ & B1 & B2 & B3 & B4 & METEOR & ROUGE$_L$ & CIDEr-D & Training Time (s) \\
\midrule
\multirow{4}{*}{LEVIR-MCI-Trees}
 & RegNetX-400MF & 85.14 & 94.81 & 75.46 & 73.35 & 57.73 & 43.32 & 32.11 & \underline{23.38} & \underline{49.40} & 46.18 & 1024 $\pm$ 3.74 \\
 & MiT-B0 & 87.12 & 95.52 & 78.72 & 72.09 & 57.76 & 42.68 & 31.03 & 22.73 & 48.98 & 42.63 & 592 $\pm$ 2.52 \\
 & MiT-B1 & \underline{88.13} & \underline{95.89} & \underline{80.36} & \textbf{75.25} & \textbf{60.90} & \textbf{46.21} & \textbf{34.41} & 23.32 & 49.34 & \textbf{48.69} & 1437 $\pm$ 27.39 \\
 & MiT-B2 & \textbf{88.62} & \textbf{96.03} & \textbf{81.20} & \underline{74.67} & \underline{60.63} & \underline{45.90} & \underline{34.10} & \textbf{23.61} & \textbf{49.83} & \underline{48.31} & 2910 $\pm$ 3.61 \\
\cmidrule(l){1-13}
\multirow{4}{*}{Forest-Change} 
 & RegNetX-400MF & 63.01 & 95.56 & 30.47 & \textbf{69.23} & \textbf{58.48} & \textbf{49.70} & \underline{42.84} & \textbf{31.57} & \underline{51.51} & \textbf{48.10} & 175 $\pm$ 4.64 \\
 & MiT-B0 & 65.86 & \underline{96.16} & 35.55 & 61.17 & 50.40 & 41.46 & 34.56 & 26.93 & 46.76 & 27.04 & 256 $\pm$ 1.00 \\
 & MiT-B1 & \underline{67.10} & 96.12 & \underline{38.07} & \underline{67.54} & 56.35 & 47.55 & 40.17 & 28.22 & 48.52 & 38.79 & 419 $\pm$ 1.53 \\
 & MiT-B2 & \textbf{68.01} & \textbf{96.35} & \textbf{39.67} & 67.25 & \underline{57.44} & \underline{49.63} & \textbf{43.23} & \underline{30.19} & \textbf{51.94} & \underline{44.15} & 986 $\pm$ 12.50 
 \\
\bottomrule
\end{tabular}
\end{table*}

However, performance gains begin to diminish beyond the MiT-B1 backbone. Improvements from MiT-B1 to MiT-B2 are relatively small, and captioning metrics show little benefit from the larger backbone, suggesting that the additional representational capacity predominantly benefits the change detection task. The lightweight RegNetX-400MF backbone produces the lowest change detection performance, especially for change class discrimination, while demonstrating notably strong captioning performance. On LEVIR-MCI-Trees, RegNetX-400MF outperforms MiT-B0 across most metrics, while on Forest-Change it achieves some of the strongest captioning scores across several metrics, including higher BLEU-1 and CIDEr-D scores than the MiT-B2 backbone. This suggests that the semantic information required for caption generation can still be captured effectively with a compact convolutional feature extractor when supported by the cross-modal reasoning components of the architecture. In contrast, improvements in backbone capacity primarily benefit the spatial localisation required for accurate change detection.

Training times also grow exponentially as backbone size increases, limiting the practicality of configurations beyond MiT-B2. While larger backbones offer modest detection improvements, the gains diminish relative to the associated computational cost. Despite slower inference times, FC-Supervised with the MiT-B2 backbone trains at a speed comparable to the benchmark models from Section \ref{section:experiments-mci-model-performance}, highlighting the efficiency of the architecture’s dual-training approach. Overall, these results indicate that backbone selection should balance representational capacity and computational efficiency, prioritising architectures that extract features useful to both change detection and captioning tasks while maintaining practical training requirements. 

\subsubsection{Forest-Chat as a Tool for Forest Change Analysis}
\label{section:experiments-forest-chat-analysis}

\begin{figure*}[ht!]
    \centering
    \includegraphics[width=0.90\linewidth]{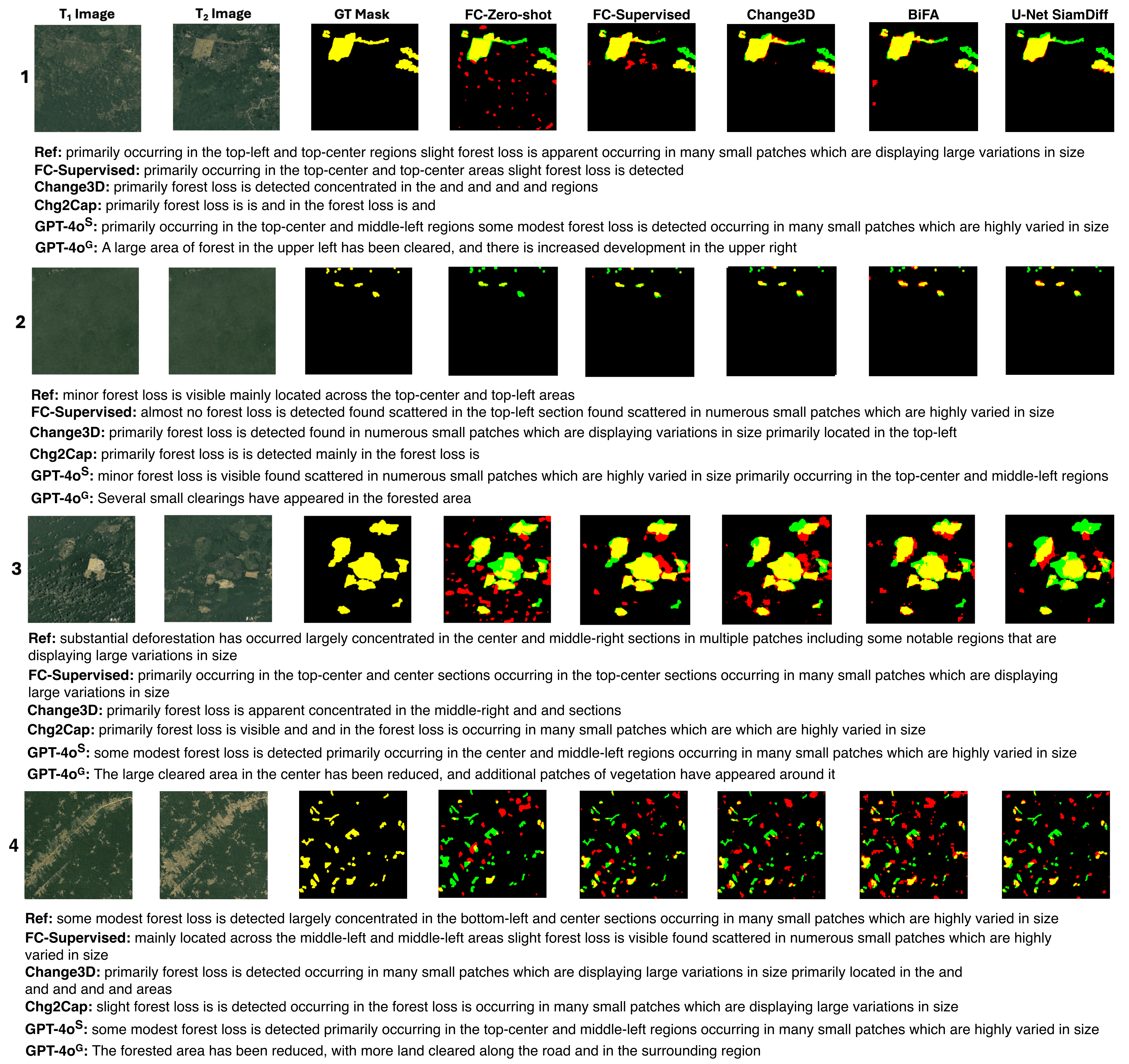}
    \caption[Qualitative Performance Examples]{A selection of qualitative comparison results between the benchmarked models on the Forest-Change test set. Yellow indicates agreement with GT mask, red for false positives, green for false negatives.}
    \label{fig:qualitative_performance}
\end{figure*}

To qualitatively assess benchmark model performance on the Forest-Change dataset, Figure \ref{fig:qualitative_performance} presents representative examples of change mask and change caption predictions from each model capable of the respective task. For change mask generation, all change detection capable methods are capable of capturing larger and medium change areas, with the FC-Supervised the most accurate for smaller change regions. All models struggle with accurately predicting numerous small, scattered change regions. This is likely attributable to the underrepresentation of small-patch examples during training, where the loss signal is dominated by the larger no-change class and small changed objects contribute disproportionately few pixels to overall mIoU. Incorporating auxiliary data such as canopy height maps could provide additional feature channels to help distinguish small change regions from surrounding vegetation, while architectural components designed for fine-grained localisation and boundary refinement represent a further avenue for mitigating these failure modes. When used in a zero-shot capacity, Forest-Chat is notably more susceptible to producing false positives, with extreme sensitivity to atmospheric noise, and often has little overlap when encountering small scattered loss patterns. Smoother change boundaries and spatial coherence are nonetheless observed for both supervised and zero-shot settings compared to the benchmarked methods, likely due to attention-based feature aggregation and region-level decoding.

For change captioning, both FC-Supervised and FC-Zero-shot reliably construct sentences capturing change severity, location, and patch characteristics. FC-Supervised is most consistent in capturing severity, though quality can occasionally degrade when describing locality and patchiness, as seen in examples 1 and 4. FC-Zero-shot constructs semantically valuable descriptions for both prompt types, but is considerably more aligned with target captions under style-guided prompting. Both Change3D and Chg2Cap routinely exhibit issues with semantic detail and sentence composition. Due to training objectives optimising against BLEU-4, all supervised models tend to replicate the rule-based phrasing present in the majority of training captions, largely ignoring the richer semantic language of human-written examples. All models are constrained by the limited linguistic diversity of training captions; richer, geographically grounded annotations would directly improve both lexical variety and spatial specificity. Common failure cases can be partially resolved through FC-Zero-shot refinement, which enriches captions with domain-aware semantics, although supervised-level performance is not reached even with style-guided prompting.

\begin{figure*}[ht!]
    \centering
    \includegraphics[width=0.80\linewidth]{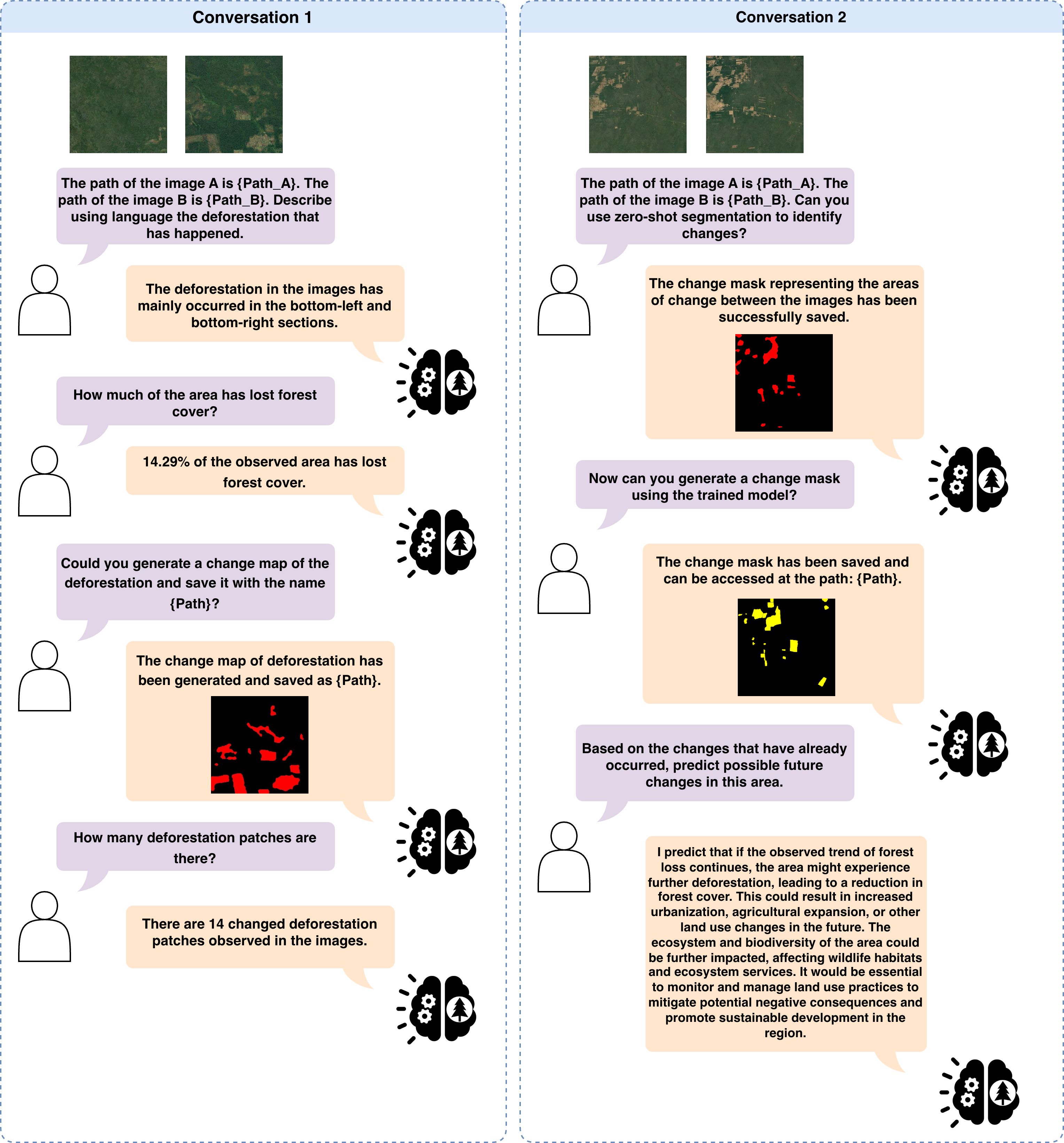}
    \caption[Forest-Chat Conversation Examples]{Conversation examples between users and Forest-Chat. Forest-Chat is capable of understanding users and orchestrating relevant tools to answer questions relating to forest change interpretation.}
    \label{fig:conversation_examples}
\end{figure*}

Figure \ref{fig:conversation_examples} presents two representative conversations with the Forest-Chat agent, demonstrating its ability to respond to user queries through natural language across a range of tasks including supervised and zero-shot change detection, change captioning, deforestation area estimation, patch counting, and future change reasoning. The agent delivers this range of capabilities through a combination of its innate knowledge and the provided toolset.

The overall toolset available to the agent is intentionally compact, with new functionality requiring manual incorporation alongside few-shot examples 
to guide tool usage awareness. Although the agent periodically fails to construct valid executable code or misunderstands aspects of a user's request, internal validation and retry mechanisms prevent erroneous responses from reaching the user. Logic failures can occur when few-shot examples do not adequately document the range of supported tasks. Exploring prompt engineering optimisation or task-aware retrieval strategies that can access pre-defined task solutions and forest ecology knowledge could significantly improve toolset utilisation and conversational richness \citep{xu2024rs}.

\section{Discussion and Future Work}
\label{section:discussion}
The development of Forest-Chat demonstrates the feasibility of adapting VLMs with conversational interfaces for forest change analysis. The system can generate both pixel-level change masks and semantic captions that allow users to explore temporal dynamics interactively, effectively identifying dominant and visually salient change patterns. At the same time, the results highlight two important remaining challenges: detecting small, spatially fragmented forest change patches, and identifying visually subtle but ecologically significant changes. From a practical deployment perspective, FC-Supervised's inference times as reported in Table \ref{tab:inference-time} are not prohibitive, with training feasible on consumer-grade CPU hardware for smaller dataset sizes. The system's end-to-end query response times are well-suited to an analytical workflow where interpretability takes precedence over real-time interaction. Where inference efficiency is a priority, training FC-Supervised or fine-tuning from the provided pretrained weights offers a more computationally efficient and higher-performing alternative to FC-Zero-shot change detection, as cross-domain transfer experiments in Section \ref{section:domain-transfer} demonstrate that only limited domain samples are required. Taken together, these findings underscore the broader challenge of deploying RSICI systems beyond urban and general-purpose domains, with Forest-Chat serving as an empirical study of the considerations involved in such adaptation.

Compared with other datasets for RSCDC, Forest-Change exhibits notable limitations in terms of scale and diversity. The RS imagery and change masks are derived from a limited number of unique sites \citep{hewarathna2024change}, and domain transfer experiments on JL1-CD-Trees further highlight the challenges of generalising across datasets with substantial resolution, seasonal, and sensor differences. Additionally, the generation of the majority of captions via a programmatic approach, compounded by limited mask diversity, resulted in suboptimal caption quality. This introduced an artificial limitation in linguistic complexity, with captions largely ignoring each scene's geographical context. We partially alleviate these issues through zero-shot caption refinement, which can enrich captions with domain-aware semantics and spatial grounding. Nonetheless, developing a large, spatially dense, and diverse forest change dataset with high-quality annotations remains crucial, particularly for capturing subtle, localised, and ecologically meaningful change processes. Integrating multiple modalities and various temporal sequence lengths would increase the variety of downstream tasks models could learn from the data.

Designing agentic VLM frameworks capable of providing further support for forest change analysis tasks is another important avenue. The current framework is limited to tasks auxiliary to change detection, captioning, and general reasoning. Enhanced visualisation tools and regression outputs could be applied for advanced ecological health assessments and biodiversity monitoring. If pursued through a single multi-task architecture, appropriate model structures and training strategies would need careful consideration of task balance and inter-task relationships. Future work should aim to develop architectures that can reliably capture small-scale forest changes informed by ecological dynamics and domain-specific knowledge.

The reliance on prompt engineering with few-shot examples for agent execution logic periodically results in the LLM misinterpreting user instructions, subsequently failing to produce a useful solution, and is inefficient for scaling to a wide range of complex scenarios. Expanding the agent's toolset, introducing forest ecology domain knowledge, integrating task-aware retrieval based on historically successful responses, and supporting a multi-agent framework or Mixture-of-Experts system would markedly enhance robustness across a broader range of operational scenarios. Systematic evaluation methodologies for LLM-driven RS agents remain an open research challenge, with recent work such as ThinkGeo \citep{shabbir2025thinkgeo} beginning to explore benchmarking frameworks for geospatial LLM system performance and capabilities.

\section{Conclusion}
\label{section:conclusion}
Forest-Chat addresses the growing need for interpretable and accessible forest change analysis by integrating VLMs and MLLMs with a conversational interface. The system leverages the MCI model alongside AnyChange and GPT-4o to provide both supervised and zero-shot pixel-level change detection and semantic captioning, enabling comprehensive bi-temporal forest change analysis. The Forest-Change dataset, pairing change masks with descriptive captions, supports joint training for detection and captioning, while multi-task loss balancing ensures effective optimisation across objectives. Supplementary tools, including a bi-temporal image captioning interface and a zero-shot point query interface, further facilitate interactive exploration and annotation, enhancing usability for researchers and practitioners.

Extensive experiments demonstrate that Forest-Chat reliably detects and describes forest changes, with the MCI model generalising well from urban to forest change contexts. Cross-domain experiments on JL1-CD-Trees highlight the challenges of transferring models across datasets with substantial resolution and seasonal differences, while demonstrating that modest target-domain supervision is sufficient to achieve effective adaptation to new forest environments. Zero-shot MLLM captioning experiments show that general prompting alone is insufficient for producing domain-consistent change descriptions, although style-guided prompting improves structural consistency. Targeted refinement can correct specific captioning errors and inject missing contextual detail as a post-processing step for supervised captioning models. 

Beyond system-level contributions, this work presents an empirical study of the challenges and considerations involved in adapting VLM-based agents to forest change analysis - a domain whose ecological diversity and lack of annotated data present non-trivial barriers beyond those encountered in urban RS tasks. Incorporating an LLM with vision-language capabilities lowers the barriers for engaging with RS data, highlighting the broader potential of interactive AI systems to improve interpretability, accessibility, and efficiency in environmental monitoring. This framework provides a foundation for future work, including scaling to larger and more diverse datasets, integrating domain knowledge and additional modalities, and supporting richer natural language queries and tasks. Together, these directions advance the development of intelligent, interpretable tools for ecological research and large-scale forest monitoring.

\section*{CRediT authorship contribution statement}
\textbf{James Brock:} Conceptualisation, Data curation, Formal analysis; Investigation, Methodology, Project administration, Software, Validation, Visualisation, Writing – original draft, Writing – review and editing. \textbf{Ce Zhang:} Supervision, Conceptualisation, Project administration, Writing – review and editing. \textbf{Nantheera Anantrasirichai:} Supervision, Writing – review and editing.

\section*{Declaration of competing interest}
The authors declare that they have no known competing financial interests or personal relationships that could have appeared to influence the work reported in this paper.

\section*{Funding}
This research did not receive any specific grant from funding agencies in the public, commercial, or not-for-profit sectors.

\section*{Acknowledgements}
The authors wish to acknowledge and thank the financial support of the UK Research and Innovation (UKRI) [Grant ref EP/Y030796/1] and the University of Bristol. We would also like to thank colleagues and reviewers for their helpful feedback and discussions relating to this work. 

\section*{Data availability}
All data and code used in this work are available in the following GitHub repository: \href{https://github.com/JamesBrockUoB/ForestChat}{ForestChat}.

\section*{Declaration of generative AI and AI-assisted technologies in the manuscript preparation process}
During the preparation of this work the authors used ChatGPT and Claude to develop and debug experimental code, produce visualisation code, and format elements within the article, such as tables and equations. After using these tools, the authors reviewed and edited the content as needed and take full responsibility for the content of the published article.

\bibliographystyle{elsarticle-harv}
\bibliography{sample}

\clearpage
\appendix
\makeatletter
\setcounter{table}{0}
\setcounter{figure}{0}
\makeatother

\section{\texorpdfstring{FC-Zero-shot Change Captioning\\Prompts}{FC-Zero-shot Change Captioning Prompts}}
\label{appendix:prompts}

All prompting modes share a common system instruction defining the model as an expert remote sensing analyst tasked with describing only genuine changes between two temporally separated aerial images.

\textbf{General Prompt}

The dataset-agnostic prompt instructs the model to produce a single concise sentence describing only observed changes between Image A (before) and Image B (after), or to output ``the scene is the same as before'' when no change is present. The prompt also enforces constraints such as avoiding references to the image identifiers and not describing unchanged content.

\textbf{Style-guided Prompts}

For datasets with characteristic caption distributions (Forest-Change and LEVIR-MCI-Trees), prompts additionally include few-shot examples and explicit stylistic rules to match the structure and vocabulary of the target dataset captions. These prompts specify:

\begin{itemize}
\item the allowed change vocabulary (e.g., forest loss, buildings, roads, trees),
\item location phrasing using compound spatial descriptors (e.g., top-center, bottom-left),
\item sentence length and grammatical style,
\item formatting constraints specific to each dataset.
\end{itemize}

Forest-Change prompts emphasise severity descriptors and patch-based descriptions of forest loss, while LEVIR-MCI-Trees prompts prioritise object-level urban changes such as building construction or road appearance.

\textbf{Refinement Prompts}

Refinement operates as a two-stage pipeline. A supervised caption produced by FC-Supervised is provided to FC-Zero-shot alongside the image pair. The prompt instructs the model to preserve the original change semantics while enriching the caption with spatial grounding and additional context observable in the imagery.

\clearpage

\begin{table}[h!]
\centering
\caption{\textbf{Selected FC-Zero-shot refinement examples on Forest-Change.} Predicted captions are produced by FC-Supervised; refined captions are generated by FC-Zero-shot given the predicted caption and bi-temporal image pair.}
\label{tab:refinement-forest}
\renewcommand{\arraystretch}{1.3}
\resizebox{\textwidth}{!}{%
\small
\begin{tabular}{p{0.35\textwidth} p{0.35\textwidth} p{0.20\textwidth}}
\toprule
\textbf{Predicted Caption} & \textbf{Refined Caption} & \textbf{Enrichment Type} \\
\midrule
primarily occurring in the top-center and top-left forest loss is detected found scattered in numerous small patches which are highly varied in size &
Primarily occurring in the top-center and top-left regions, forest loss is detected found scattered in numerous small patches which are highly varied in size, with some areas showing sharp rectangular boundaries consistent with clearcutting &
Change process inference \\
\midrule
mainly located across the top-center and top-left regions some modest forest loss is visible &
Mainly located across the top-center and top-left regions some modest forest loss is visible occurring in many small patches which are moderately varied in size with exposed understorey &
Forest cover context \\
\midrule
slight forest loss is noted mainly located across the middle-right and middle-right and middle-right sections found scattered in size mainly located across the middle-right and bottom &
Slight forest loss is noted mainly located across the middle-right and bottom sections occurring in many small patches which are highly varied in size with exposed understorey &
Spatial grounding \& artefact correction \\
\midrule
primarily occurring in the middle-left and middle-left areas slight forest loss is detected occurring in many small patches which are displaying large variations in size &
Primarily occurring in the middle-left and center areas slight forest loss is detected occurring in many small patches which are displaying large variations in size with some fragmentation along existing clearings &
Geographic feature reference \\
\bottomrule
\end{tabular}%
}
\end{table}

\begin{table}[h!]
\centering
\caption{\textbf{Selected FC-Zero-shot caption refinement examples on LEVIR-MCI-Trees.} Predicted captions are produced by FC-Supervised; refined captions are generated by FC-Zero-shot given the predicted caption and bi-temporal image pair.}
\label{tab:refinement-levir}
\renewcommand{\arraystretch}{1.3}
\resizebox{\textwidth}{!}{%
\small
\begin{tabular}{p{0.35\textwidth} p{0.35\textwidth} p{0.20\textwidth}}
\toprule
\textbf{Predicted Caption} & \textbf{Refined Caption} & \textbf{Enrichment Type} \\
\midrule
a road and a road with villas built along the road &
Some villas are built along the road in the center, replacing the dense vegetation and sparse trees . &
Spatial localisation \& land cover context \\
\midrule
a small house appears in the bareland &
A small house appears near the bottom left, and a pond emerges at the top right, replacing some vegetation . &
Secondary change detection \\
\midrule
a road with houses are built along the road &
Some houses are built around a cul-de-sac and replace the sparse vegetation alongside the road . &
Infrastructure reasoning \\
\midrule
a house appears in the woods &
A house appears beside the existing house, and some trees are removed alongside the road . &
Contextual spatial grounding \\
\bottomrule
\end{tabular}%
}
\end{table}

\clearpage

\section{Supplementary Results}
\setcounter{table}{0}

\begin{table}[h!]
\centering
\caption{\textbf{Full MTL strategy change detection and change captioning ablation results on the LEVIR-MCI-Trees dataset.} Mean $\pm$ standard deviation across configurations, with three runs each. Bold indicates the best mean value per metric within each method group. When results are tied or close, preference is given to the method with the lower standard deviation.}
\label{tab:mtl-ablation-results-levir}
\setlength{\tabcolsep}{3pt}
\resizebox{\textwidth}{!}{%
\begin{tabular}{l | c c c | c c c c c c c}
\toprule
\textbf{MTL Configuration} & mIoU & $IoU_{\textit{nc}}$ & $IoU_{\textit{c}}$ & B1 & B2 & B3 & B4 & METEOR & ROUGE$_L$ & CIDEr-D \\
\midrule
Equal & 88.13$\pm$0.16 & 95.89$\pm$0.05 & 80.36$\pm$0.27 & 75.25$\pm$0.69 & 60.90$\pm$0.54 & 46.21$\pm$0.23 & 34.41$\pm$0.19 & 23.32$\pm$0.38 & 49.34$\pm$0.34 & 48.69$\pm$1.37 \\
+ CAGrad & 88.16$\pm$0.16 & 95.89$\pm$0.06 & 80.43$\pm$0.25 & 74.15$\pm$1.31 & 59.16$\pm$1.41 & 44.26$\pm$1.02 & 32.61$\pm$1.22 & 23.48$\pm$0.60 & 49.84$\pm$1.42 & 49.47$\pm$2.42 \\
+ PCGrad & 88.13$\pm$0.07 & \textbf{95.90$\pm$0.02} & 80.36$\pm$0.15 & 75.67$\pm$0.73 & 60.28$\pm$0.87 & 45.35$\pm$0.97 & 33.55$\pm$0.73 & \textbf{24.05$\pm$0.93} & \textbf{50.35$\pm$0.90} & 48.84$\pm$3.15 \\
+ GradDrop & 87.90$\pm$0.11 & 95.84$\pm$0.04 & 79.96$\pm$0.20 & 73.57$\pm$1.69 & 59.18$\pm$1.61 & 43.88$\pm$1.40 & 31.93$\pm$0.69 & 23.07$\pm$0.46 & 49.48$\pm$0.67 & 45.21$\pm$2.98 \\
\midrule
EDWA & 87.90$\pm$0.13 & 95.77$\pm$0.03 & 80.03$\pm$0.24 & 74.78$\pm$2.49 & 59.91$\pm$1.84 & 44.56$\pm$2.32 & 32.41$\pm$2.69 & 23.39$\pm$0.86 & 49.31$\pm$0.85 & 45.84$\pm$4.18 \\
+ CAGrad & 87.04$\pm$0.43 & 95.51$\pm$0.13 & 78.56$\pm$0.76 & 70.55$\pm$5.96 & 55.88$\pm$5.86 & 40.28$\pm$6.41 & 28.62$\pm$5.42 & 21.82$\pm$2.59 & 47.43$\pm$3.08 & 40.99$\pm$8.12 \\
+ PCGrad & 86.92$\pm$0.45 & 95.41$\pm$0.13 & 78.42$\pm$0.76 & 70.38$\pm$4.30 & 56.13$\pm$3.89 & 40.55$\pm$5.31 & 29.34$\pm$4.89 & 21.60$\pm$1.79 & 47.56$\pm$1.54 & 40.06$\pm$9.77 \\
+ GradDrop & 86.81$\pm$0.40 & 95.42$\pm$0.09 & 78.21$\pm$0.72 & 73.18$\pm$3.16 & 58.82$\pm$2.54 & 43.85$\pm$3.01 & 32.03$\pm$3.78 & 22.96$\pm$0.97 & 49.04$\pm$0.89 & 45.86$\pm$7.66 \\
\midrule
Uncertainty & 87.52$\pm$0.28 & 95.65$\pm$0.09 & 79.39$\pm$0.47 & 74.36$\pm$1.11 & 59.89$\pm$0.59 & 45.32$\pm$0.65 & 33.76$\pm$1.11 & 23.29$\pm$0.78 & 49.55$\pm$0.96 & 47.78$\pm$4.21 \\
+ CAGrad & \textbf{88.20$\pm$0.09} & 95.90$\pm$0.03 & \textbf{80.51$\pm$0.16} & \textbf{75.84$\pm$0.30} & \textbf{61.25$\pm$0.61} & \textbf{47.25$\pm$0.60} & \textbf{36.08$\pm$0.45} & 23.72$\pm$0.41 & 50.27$\pm$0.71 & \textbf{51.82$\pm$3.09} \\
+ PCGrad & 87.09$\pm$0.16 & 95.48$\pm$0.09 & 78.70$\pm$0.23 & 73.67$\pm$1.85 & 58.50$\pm$1.28 & 43.68$\pm$0.92 & 31.84$\pm$0.91 & 23.09$\pm$0.70 & 49.40$\pm$1.06 & 47.55$\pm$0.55 \\
+ GradDrop & 87.22$\pm$0.23 & 95.54$\pm$0.10 & 78.91$\pm$0.38 & 75.01$\pm$1.29 & 60.90$\pm$1.16 & 45.58$\pm$1.41 & 33.61$\pm$1.68 & 23.55$\pm$0.11 & 50.33$\pm$0.21 & 48.07$\pm$2.82 \\
\bottomrule
\end{tabular}%
}
\end{table}

\begin{table}[h!]
\centering
\caption{\textbf{Full MTL change detection and captioning ablation results on the LEVIR-MCI-Trees dataset (percentage changes relative to baseline).} Only $\Delta\%$ values are reported; positive/negative values indicate improvement or decline compared to the baseline configuration respectively. Bold indicates the largest improvement among non-baseline configurations per metric when the improvement is clearly distinguishable.}
\label{tab:mtl-ablation-delta-levir}
\resizebox{\textwidth}{!}{%
\begin{tabular}{l | c c c | c c c c c c c}
\toprule
\textbf{MTL Configuration} & mIoU & IoU$_{\textit{nc}}$ & IoU$_{\textit{c}}$ & B1 & B2 & B3 & B4 & METEOR & ROUGE$_L$ & CIDEr-D \\
\midrule
Baseline (Equal) & - & - & - & - & - & - & - & - & - & - \\
+ CAGrad & +0.03 & +0.00 & +0.08 & -1.47 & -2.86 & -4.21 & -5.23 & +0.67 & +1.01 & +1.60 \\
+ PCGrad & +0.00 & \textbf{+0.01} & +0.00 & +0.56 & -1.02 & -1.86 & -2.48 & \textbf{+3.12} & \textbf{+2.05} & +0.32 \\
+ GradDrop & -0.26 & -0.05 & -0.50 & -2.24 & -2.83 & -5.04 & -7.20 & -1.08 & +0.28 & -7.15 \\
\midrule
EDWA & -0.26 & -0.12 & -0.41 & -0.62 & -1.62 & -3.57 & -5.82 & +0.29 & -0.06 & -5.85 \\
+ CAGrad & -1.24 & -0.40 & -2.25 & -6.25 & -8.24 & -12.84 & -16.83 & -6.45 & -3.86 & -15.81 \\
+ PCGrad & -1.38 & -0.50 & -2.41 & -6.47 & -7.84 & -12.23 & -14.74 & -7.38 & -3.61 & -17.73 \\
+ GradDrop & -1.50 & -0.49 & -2.67 & -2.75 & -3.41 & -5.10 & -6.92 & -1.57 & -0.61 & -5.81 \\
\midrule
Uncertainty & -0.69 & -0.24 & -1.21 & -1.18 & -1.66 & -1.92 & -1.89 & -0.15 & +0.43 & -1.87 \\
+ CAGrad & \textbf{+0.08} & \textbf{+0.01} & \textbf{+0.19} & \textbf{+0.78} & \textbf{+0.57} & \textbf{+2.26} & \textbf{+4.86} & +1.71 & +1.88 & \textbf{+6.43} \\
+ PCGrad & -1.18 & -0.43 & -2.07 & -2.10 & -3.95 & -5.48 & -7.49 & -0.99 & +0.13 & -2.34 \\
+ GradDrop & -1.03 & -0.37 & -1.81 & -0.32 & +0.00 & -1.36 & -2.28 & +0.98 & +2.00 & -1.27 \\
\bottomrule
\end{tabular}%
}
\end{table}

\clearpage

\begin{table}[h!]
\centering
\caption{\textbf{Full MTL strategy change detection and change captioning ablation results on the Forest-Change dataset.} Mean $\pm$ standard deviation across configurations, with three runs each. Bold indicates the best mean value per metric within each method group. When results are tied or close, preference is given to the method with the lower standard deviation.}
\label{tab:mtl-ablation-results-fc}
\setlength{\tabcolsep}{3pt}
\resizebox{\textwidth}{!}{%
\begin{tabular}{l | c c c | c c c c c c c}
\toprule
\textbf{MTL Configuration} & mIoU & $IoU_{\textit{nc}}$ & $IoU_{\textit{c}}$ & B1 & B2 & B3 & B4 & METEOR & ROUGE$_L$ & CIDEr-D \\
\midrule
Equal
& 67.10$\pm$0.07 & 96.12$\pm$0.09 & 38.07$\pm$0.09
& \textbf{67.88$\pm$4.55} & \textbf{56.35$\pm$4.88} & 47.55$\pm$5.31 & 40.17$\pm$5.84
& 28.22$\pm$2.61 & 48.52$\pm$3.46 & \textbf{38.79$\pm$10.90} \\
+ CAGrad
& 67.02$\pm$0.30 & 96.29$\pm$0.15 & 37.75$\pm$0.45
& 65.09$\pm$3.51 & 55.49$\pm$3.41 & \textbf{47.87$\pm$4.04} & \textbf{41.55$\pm$4.46}
& 28.28$\pm$1.17 & 49.39$\pm$1.20 & 38.49$\pm$5.62 \\
+ PCGrad
& 67.46$\pm$0.38 & 96.18$\pm$0.11 & 38.74$\pm$0.90
& 64.28$\pm$3.26 & 52.92$\pm$3.29 & 44.22$\pm$3.44 & 37.56$\pm$1.96
& 27.46$\pm$1.25 & 46.72$\pm$1.43 & 34.72$\pm$6.66 \\
+ GradDrop
& \textbf{67.77$\pm$0.46} & 96.21$\pm$0.10 & \textbf{39.33$\pm$0.81}
& 61.12$\pm$3.83 & 50.77$\pm$4.25 & 41.67$\pm$4.32 & 34.44$\pm$3.56
& 26.25$\pm$2.20 & 45.56$\pm$2.78 & 30.29$\pm$8.85 \\
\midrule
EDWA
& 67.30$\pm$0.49 & 96.20$\pm$0.25 & 38.40$\pm$0.92
& 66.16$\pm$4.91 & 55.54$\pm$5.12 & 46.42$\pm$4.14 & 39.79$\pm$3.68
& 28.40$\pm$3.54 & 49.02$\pm$3.48 & 36.16$\pm$7.21 \\
+ CAGrad
& 66.98$\pm$0.73 & 96.16$\pm$0.22 & 37.79$\pm$1.27
& 63.30$\pm$1.91 & 52.17$\pm$3.57 & 43.12$\pm$4.28 & 35.85$\pm$4.83
& 26.73$\pm$0.40 & 46.59$\pm$1.52 & 29.96$\pm$4.05 \\
+ PCGrad
& 66.32$\pm$0.68 & 96.13$\pm$0.12 & 36.52$\pm$1.51
& 65.84$\pm$3.61 & 55.36$\pm$2.87 & 46.55$\pm$2.63 & 39.64$\pm$2.84
& 27.53$\pm$2.13 & 47.54$\pm$2.11 & 35.70$\pm$4.88 \\
+ GradDrop
& 66.90$\pm$0.63 & 96.26$\pm$0.19 & 37.54$\pm$1.55
& 62.52$\pm$6.11 & 51.68$\pm$5.90 & 42.71$\pm$5.17 & 35.63$\pm$5.51
& 27.21$\pm$4.19 & 47.69$\pm$4.98 & 30.69$\pm$6.13 \\
\midrule
Uncertainty
& 67.30$\pm$0.23 & \textbf{96.41$\pm$0.20} & 38.19$\pm$0.44
& 64.29$\pm$2.76 & 54.64$\pm$2.45 & 46.66$\pm$2.42 & 40.19$\pm$2.64
& 28.41$\pm$2.06 & 48.67$\pm$2.66 & 35.58$\pm$11.46 \\
+ CAGrad
& 67.49$\pm$0.49 & 96.35$\pm$0.06 & 38.64$\pm$0.92
& 62.64$\pm$1.99 & 52.83$\pm$1.24 & 45.43$\pm$0.27 & 39.61$\pm$0.62
& 28.31$\pm$1.45 & 49.22$\pm$2.11 & 36.71$\pm$5.89 \\
+ PCGrad
& 66.82$\pm$0.48 & 96.26$\pm$0.17 & 37.37$\pm$0.81
& 63.04$\pm$4.75 & 52.89$\pm$5.14 & 44.07$\pm$5.74 & 37.20$\pm$6.42
& 26.77$\pm$3.24 & 47.18$\pm$4.32 & 34.68$\pm$2.66 \\
+ GradDrop
& 66.48$\pm$0.38 & 95.86$\pm$0.43 & 37.11$\pm$0.55
& 65.57$\pm$4.09 & 55.18$\pm$4.93 & 46.42$\pm$5.05 & 39.39$\pm$5.27
& 28.29$\pm$2.29 & \textbf{49.61$\pm$3.08} & 27.97$\pm$5.81 \\
\bottomrule
\end{tabular}%
}
\end{table}

\begin{table}[h!]
\centering
\caption{\textbf{Full MTL change detection and captioning ablation results on the Forest-Change dataset, shown as percentage changes relative to baseline.} Only $\Delta\%$ values are reported; positive/negative values indicate improvement or decline compared to the baseline configuration respectively. Bold indicates the largest improvement among non-baseline configurations per metric when the improvement is clearly distinguishable.}
\label{tab:mtl-ablation-delta-fc}
\resizebox{\textwidth}{!}{%
\begin{tabular}{l | c c c | c c c c c c c}
\toprule
\textbf{MTL Configuration} & mIoU & IoU$_{\textit{nc}}$ & IoU$_{\textit{c}}$ & B1 & B2 & B3 & B4 & METEOR & ROUGE$_L$ & CIDEr-D \\
\midrule
Baseline (Equal) & – & – & – & – & – & – & – & – & – & – \\
+ CAGrad & -0.12 & +0.18 & -0.84 & -4.11 & -1.53 & \textbf{+0.67} & \textbf{+3.44} & +0.21 & +1.79 & -0.77 \\
+ PCGrad & +0.54 & +0.06 & +1.76 & -5.30 & -6.09 & -7.00 & -6.50 & -2.69 & -3.71 & -10.49 \\
+ GradDrop & \textbf{+1.00} & +0.09 & \textbf{+3.31} & -9.96 & -9.90 & -12.37 & -14.26 & -6.98 & -6.10 & -21.91 \\
\midrule
EDWA & +0.30 & +0.08 & +0.87 & -2.53 & -1.44 & -2.38 & -0.95 & +0.64 & +1.03 & -6.78 \\
+ CAGrad & -0.18 & +0.04 & -0.74 & -6.75 & -7.42 & -9.32 & -10.75 & -5.28 & -3.98 & -22.76 \\
+ PCGrad & -1.16 & +0.01 & -4.07 & -3.01 & -1.76 & -2.10 & -1.32 & -2.45 & -2.02 & -7.97 \\
+ GradDrop & -0.30 & +0.15 & -1.39 & -7.90 & -8.29 & -10.18 & -11.30 & -3.58 & -1.71 & -20.88 \\
\midrule
Uncertainty & +0.30 & \textbf{+0.30} & +0.32 & -5.29 & -3.03 & -1.87 & +0.05 & +0.67 & +0.31 & -8.28 \\
+ CAGrad & +0.58 & +0.24 & +1.50 & -7.72 & -6.25 & -4.46 & -1.39 & +0.32 & +1.44 & -5.36 \\
+ PCGrad & -0.42 & +0.15 & -1.84 & -7.13 & -6.14 & -7.32 & -7.39 & -5.14 & -2.76 & -10.60 \\
+ GradDrop & -0.92 & -0.27 & -2.52 & -3.40 & -2.08 & -2.38 & -1.94 & +0.25 & \textbf{+2.25} & -27.89 \\
\bottomrule
\end{tabular}%
}
\end{table}

\clearpage

\begin{figure*}[hbt!]
\centering
\includegraphics[width=0.65\textwidth]{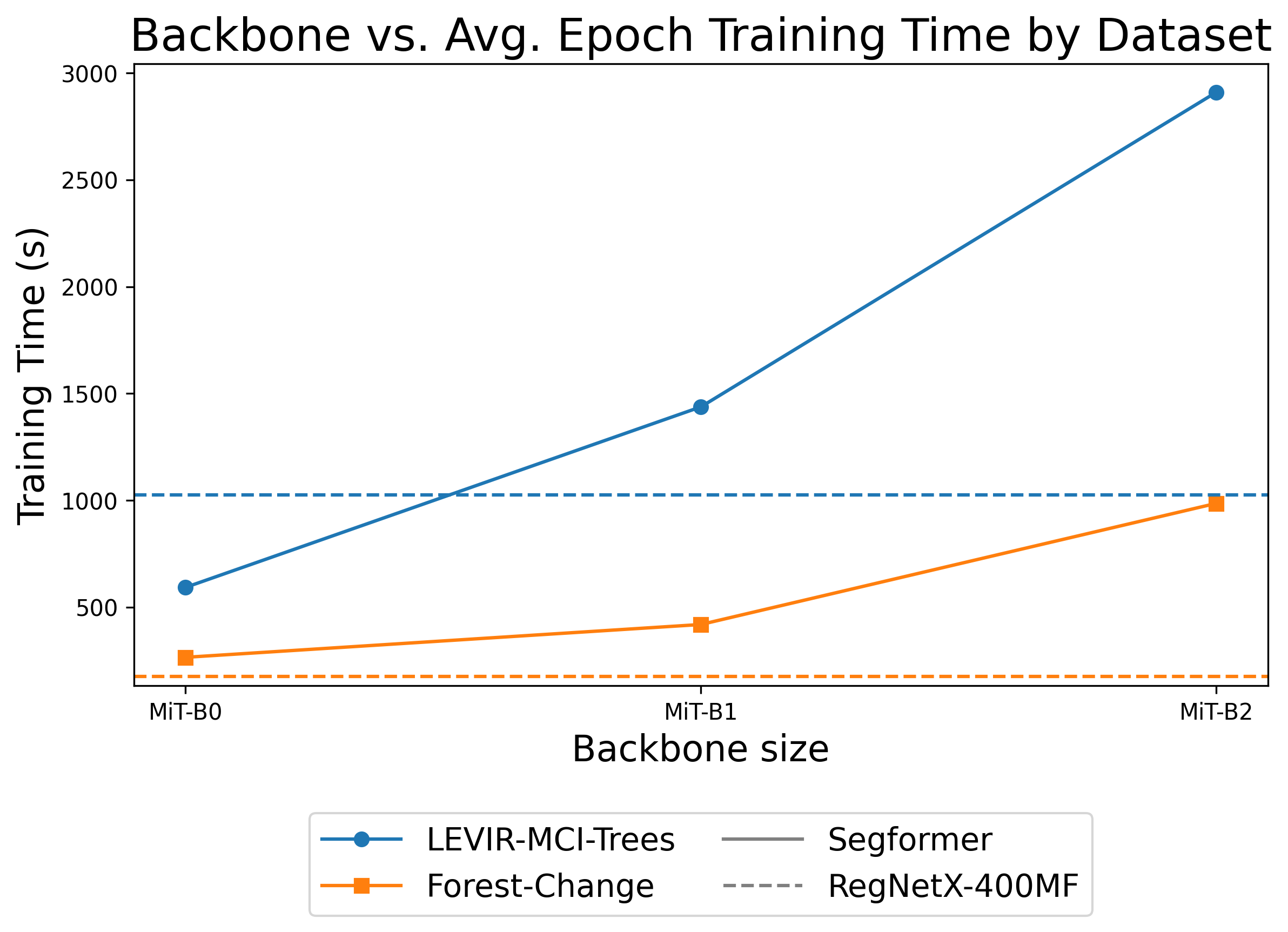}
\caption{Training time comparison by backbone size per dataset. The lines show the training time for one epoch (mean over three runs) per backbone size for the LEVIR-MCI-Trees and Forest-Change datasets. Segformer backbones are defined on the x-axis, whilst the training times for RegNetX-400MF are given along the y-axis.}
\label{fig:training-times-backbone}
\end{figure*}

\end{document}